\documentclass[journal]{IEEEtranTIE}
\pdfoutput=1

\usepackage{amssymb, subfigure, setspace, multirow}
\usepackage{verbatim}
\usepackage{authblk}
\usepackage{amsmath}
\usepackage{hhline}
\usepackage{cite}
\usepackage{algorithmic}
\usepackage{algorithm}
\usepackage{array}
\usepackage{url}
\usepackage{bm}
\usepackage{mathrsfs}
\usepackage{amsfonts}
\usepackage{graphicx}
\usepackage{textcomp}
\usepackage{xcolor}
\usepackage{microtype}
\usepackage{gensymb}
\usepackage{balance}
\usepackage{hyperref}

\newcommand{\ie}{{\em i.e.}}
\newcommand{\eg}{{\em e.g.}}
\newcommand{\et}{{\em et al.}}

\def\spth{\textsuperscript{th}}

\begin{document}
\title{WiFi-Inertial Indoor Pose Estimation for Micro Aerial Vehicles}

\author{Shengkai~Zhang,~\IEEEmembership{Student~Member,~IEEE, }
		Wei~Wang,~\IEEEmembership{Senior~Member,~IEEE, }
        and~Tao~Jiang,~\IEEEmembership{Fellow,~IEEE}
 \thanks{This work was supported in part by the National Key R\&D Program of China under Grant 2019YFB180003400, Young Elite Scientists Sponsorship Program by CAST under Grant 2018QNRC001, National Science Foundation of China with Grant 91738202.}
\thanks{S. Zhang, W. Wang, and T. Jiang are with the School of Electronic Information and Communications, Huazhong University of Science and Technology, Wuhan, China. E-mail: \{\href{mailto:szhangk@hust.edu.cn}{szhangk}, \href{mailto:weiwangw@hust.edu.cn}{weiwangw}, \href{mailto:taojiang@hust.edu.cn}{taojiang}\}@hust.edu.cn.}
}

\maketitle

\begin{abstract}
This paper presents an indoor pose estimation system for micro aerial vehicles (MAVs) with a single WiFi access point. Conventional approaches based on computer vision are limited by illumination conditions and environmental texture. Our system is free of visual limitations and instantly deployable, working upon existing WiFi infrastructure without any deployment cost. Our system consists of two coupled modules. First, we propose an angle-of-arrival (AoA) estimation algorithm to estimate MAV attitudes and disentangle the AoA for positioning. Second, we formulate a WiFi-inertial sensor fusion model that fuses the AoA and the odometry measured by inertial sensors to optimize MAV poses. Considering the practicality of MAVs, our system is designed to be real-time and initialization-free for the need of agile flight in unknown environments. The indoor experiments show that our system achieves the accuracy of pose estimation with the position error of $61.7$ cm and the attitude error of $0.92\degree$. 
\end{abstract}


\markboth{IEEE TRANSACTIONS ON INDUSTRIAL ELECTRONICS}
{}

\definecolor{limegreen}{rgb}{0.2, 0.8, 0.2}
\definecolor{forestgreen}{rgb}{0.13, 0.55, 0.13}
\definecolor{greenhtml}{rgb}{0.0, 0.5, 0.0}

\section{Introduction}
\label{sec:intro}
\IEEEPARstart{M}{icro} aerial vehicles (MAVs) are great robotic platforms for a wide range of applications in indoors such as warehouse inventory, search and rescue, and hazard detection~\cite{walmart, lin2019kalman, xiao2019sensor, he2019state}, thanks to their low cost, small size, and agile mobility. Autonomous flight is essential to these application. For example, Walmart has decided to use MAVs to replace the jobs of inventory quality assurance employees, cutting inventory checks across massive distribution centers from one month down to a single day~\cite{walmart}. These vehicles are required to autonomously check inventory in confined indoor areas.

A fundamental problem for indoor autonomous operations of MAVs is the pose estimation~\cite{islam2018observer}. The {\em pose}, including position and attitude, is the key to the flight control system of an aerial vehicle that adjusts the rotating speed of rotors to achieve desired actions for responding remote control or autonomous operations. While computer vision (CV) systems can perform accurate pose estimation, their performance suffers from illumination conditions and environmental texture~\cite{lin2018autonomous, fu2017robust, tang2019vision}. In particular, most of warehouses lack proper lighting because of their high ceilings, stacks shading, and energy conservation~\cite{fichtinger2015assessing}. This may fail to capture enough visual features and degrade the accuracy of pose estimation. 

RF-based tracking and localization offers an alternative sensing modality that is highly robust to visual limitations. Motivated by recent advances of RF-based localization~\cite{kotaru2015spotfi, luo2019dynamic}, in this paper, we aim to build an accurate RF-based pose estimator for MAVs. Such a system desires three goals:
\begin{itemize}
	\item {\bf Lightweight}: Due to the small size and limited battery capacity of MAVs, the system should be lightweight that works without adding extra sensors.
	\item {\bf Real-time}: The pose estimation should be real-time due to the fast dynamics of MAVs. 
	\item {\bf Deployable}: MAVs favor a pose estimator that is instantly deployable so that they are operable immediately in unknown environments.
\end{itemize}
Although there have been many types of RF-based localization systems such as UWB and RFID~\cite{mueller2015fusing, saab2016novel}, they cannot realize all three goals for pose estimation. UWB-based systems need extra costs to deploy specialized infrastructure that emits ultra wideband signals for ranging. RFID tags are only operable within a range of few meters, requiring to deploy dense RFID readers. Fortunately, we find that commodity WiFi offers a great opportunity to achieve these goals in that 1) it is ubiquitously available, making a system functional without deployment cost; 2) the pose estimation can be enabled upon the communication module of a MAV without extra sensors. 

\begin{figure}[t!]
	\centering
	\begin{minipage}[b]{0.24\textwidth}\centering
		\center
		\includegraphics[width=1\textwidth]{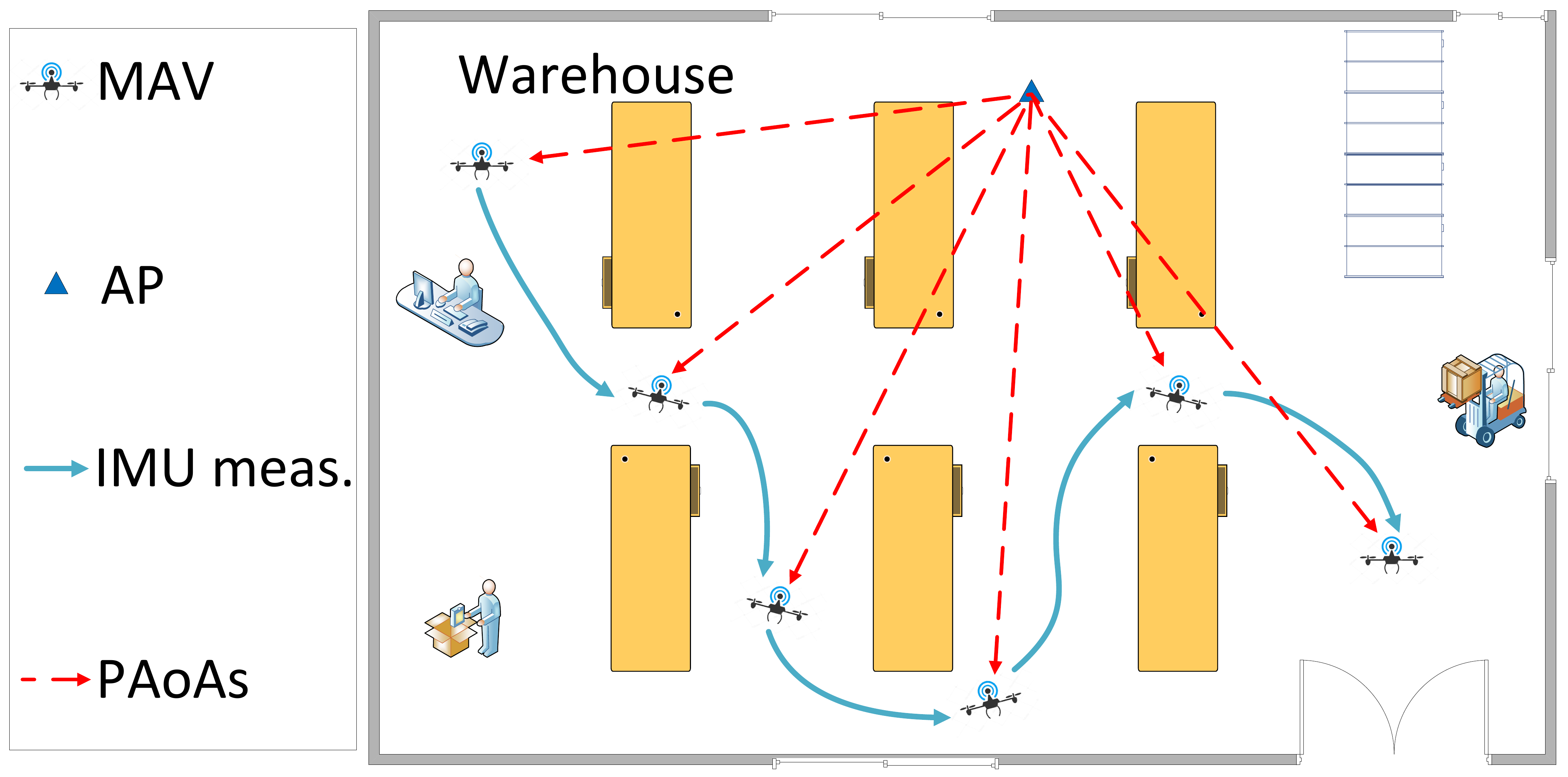}\vspace{-0.3cm}
		\caption{Use example: a MAV autonomously flies to automate inventory in a warehouse with a single commodity AP.} \label{fig:toy}
	\end{minipage}
	\hspace{-0.1cm}
	\begin{minipage}[b]{0.24\textwidth}\centering
		\center
		\includegraphics[width=1\textwidth]{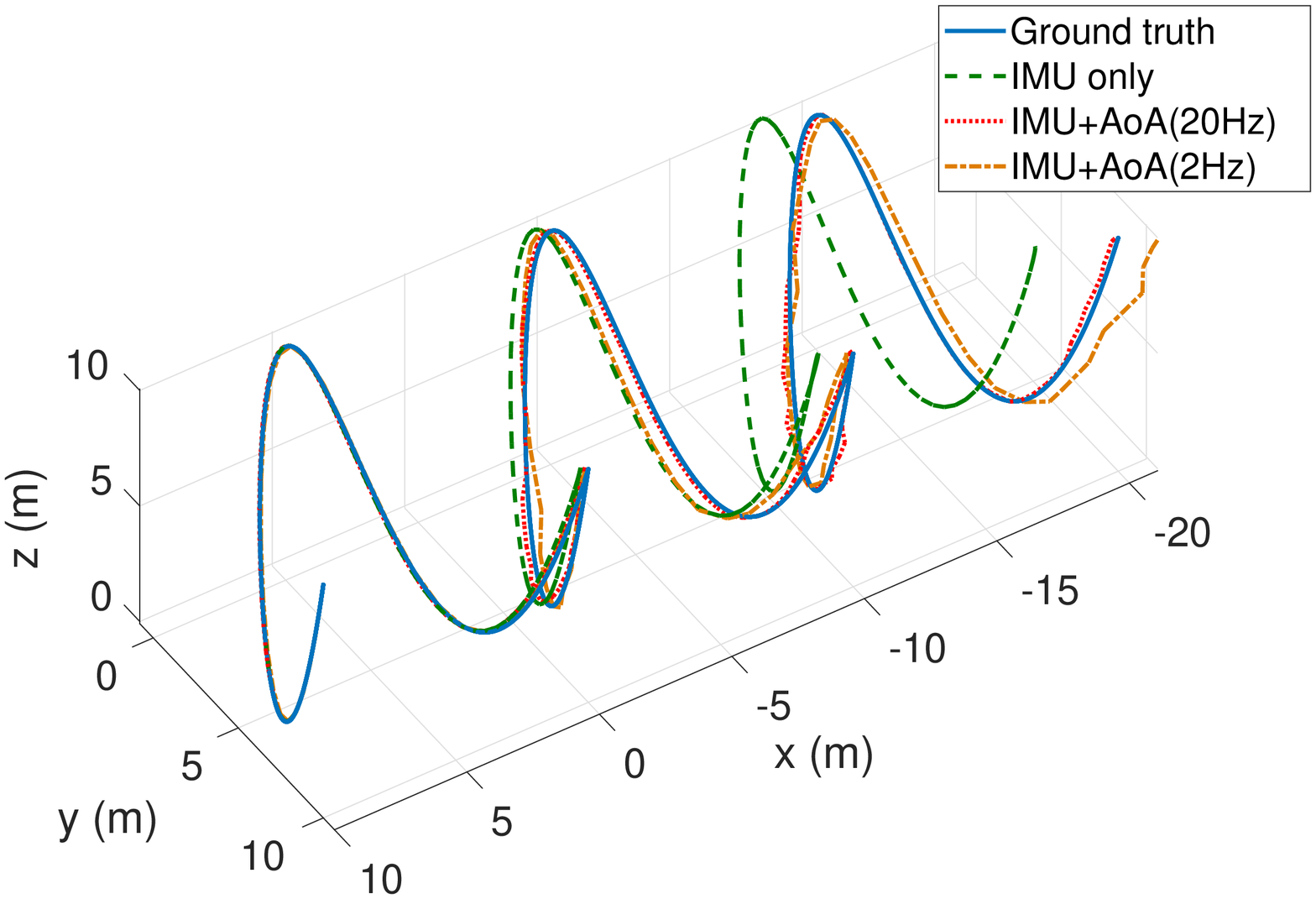}\vspace{-0.3cm}
		\caption{Simulation results: the positioning performance between integrating IMU-only measurements and AoA-IMU fusion.} \label{fig:verify}
	\end{minipage}
\end{figure}

In this paper, we present WINS, a WiFi-inertial pose estimator that allows a MAV to autonomously fly through an indoor venue assisted by a single commercial access point (AP) as shown in Fig.~\ref{fig:toy}. WINS is extremely lightweight and instantly deployable. It works upon ubiquitous WiFi infrastructure with a three-antenna linear array on the MAV, which is supported by off-the-shelf WiFi network cards. 

Despite the advantages of WINS, realizing it poses several challenges. {\em First}, with narrowband WiFi signals, one can use the angle-of-arrival (AoA\footnote{For simplicity, let us view the AoA as the angle between incident signals and the normal direction of a linear array.}) measured by an antenna array to determine positions through triangulation. Unfortunately, existing AoA-based localization approaches usually localize the target offline, computing the AoAs of deployed APs who are fixed in the venue. However, autonomous MAVs favor onboard sensing and computation for robustness~\cite{james2018liftoff} and thus the antenna array is required to be onboard to directly obtain the AoA of the MAV. In this case, the AoA can be a mixed result from the MAV's attitudes and positions. The MAV can only be localized by the partial component of the obtained AoA that contributes to positions, namely the positioning AoA (PAoA). {\em Second}, existing AoA estimation techniques have a trade-off between the accuracy and the computational cost while we desire both, being real-time and accurate, for agile flight. {\em Third}, conventional WiFi localization systems need the prior knowledge of AP deployment to define a coordinate system. However, an autonomous MAV should be able to work in unknown environments, \ie, APs' positions are unknown.

WINS addresses these challenges by two modules. We first propose a PAoA estimation algorithm that extracts the PAoA from the mixed result. The algorithm leverages the insight of the multi-view reconstruction in CV~\cite{martinec2007robust} to estimate drift-free rotations by IMU measurements and then disentangle the PAoA. In addition, the computational bottleneck of the AoA estimation lies on the AoA searching process. We utilize the odometry measured by IMU to predict an AoA candidate when receiving a WiFi packet to shrink the search scope, improving the computational speed without sacrificing the accuracy. Second, we formulate a sensor fusion model that properly takes advantages of the strengths of heterogeneous sensors, \ie, an IMU and an antenna array, to jointly optimize MAV poses. Specifically, an IMU measures odometry but suffers from a temporal drift while an antenna array provides noisy but drift-free AoAs. Fusing them together by solving a maximum likelihood problem with constraints of PAoAs and odometry is able to accurately estimate MAV poses. Considering the practicality of MAVs, the sensor fusion module is designed to be initialization-free. It can recover the initial positions of the AP and the MAV with enough motions so that WINS works without the prior knowledge of AP deployment. 

\noindent {\bf Results:} We implement WINS on the DJI M100 platform where it equips an IMU and an Intel Next Unit of Computing (NUC). The NUC uses an Intel 5300 wireless card attaching three antennas to obtain wireless channel information for AoA estimation via the 802.11 CSI tool~\cite{halperin2011tool}. The experiments are conducted in the MAV test site of our laboratory to validate individual system modules as well as the overall performance. The results show that WINS achieves the accuracy of pose estimation with the position error of $61.7$ cm and the attitude error of $0.92\degree$ via a single AP during a flight with the maximum velocity of $1.27$ m/s. 

\noindent {\bf Contributions:} WINS is a lightweight indoor pose estimator for MAVs that works upon ubiquitous WiFi via a single AP. WINS achieves this through two modules: PAoA estimation and sensor fusion. WINS properly takes advantages of heterogeneous sensors, \ie, an antenna array and an IMU, to combat the sensor noise in pose estimation. Considering the practicality of MAVs, WINS is designed to be real-time and initialization-free. We implement WINS on commodity devices and experimentally validate the system in indoors.

\section{Related Work}
\label{sec:related}
{\bf WiFi-based approaches}: Our WINS works upon existing ubiquitously available WiFi APs, offering an instantly deployable pose estimation system. J. Huang~\et~\cite{huang2011efficient} proposed WiFi-SLAM that performs simultaneous localization and mapping using the signal strength of WiFi signals in indoor environments. It is a fingerprint-based approach that can automatically label the signal strength measurements with pedometry. However, since the signal strength poses severe ambiguities due to the multipath propagation in indoors, their accuracies are meter-level.

Recent advances of wireless technologies enable the CSI extraction for WiFi OFDM signals~\cite{halperin2011tool}. Different from the signal strength, CSI provides the phase information of a signal and the channel response of multiple subcarriers, which significantly reduces the spatial ambiguity and improves the multipath resolution in indoors. CWISE~\cite{li2016csi} is the first system that leverages the CSI to estimate a MAV's position and velocity as well as the APs' positions. It pioneers the feasibility of enabling autonomous navigation via WiFi. It operates correctly in outdoors without multipath fading. The fundamental difference in our context is that wireless signals suffer severe multipath fading in indoors, making the phase information of CSI conceal the positioning information among multipath reflections. Moreover, CWISE assumes that the MAV has no rotations during the flight and requires multiple APs to address the location. In contrast, WINS designs novel algorithms to 1) estimate the MAV's attitude and disentangle the PAoA; 2) localize the MAV with a single AP; 3) combat the multipath fading to work in indoors. 

{\bf Vision/laser-based approaches}: On one hand, computer vision techniques have been used in the field of robotic autonomy in both industry and academia due to their low cost and high resolution in understanding environments. Monocular vision-based approaches~\cite{lin2018autonomous, shen2015tightly} represent the lightweight sensor suite to achieve accurate pose estimation. They are particularly suitable for small platforms, \eg, MAVs. However, their performance is sensitive to the environment visual features and lighting conditions, hindering its usage in vision-crippled areas, \eg, dim warehouses for inventory management, smoky buildings for firefighting operations. In contrast, WINS uses ubiquitous WiFi, which is highly resilient to visual limitations. 

On the other hand, laser ranging sensors, \eg, LiDAR, directly provide the metric scale and map profile of environments, providing best accuracy among existing solutions~\cite{dube2017online, hess2016real}. However, they tend to impact a MAV's lifetime because of the added weight. Moreover, the high cost (hundreds of dollars) hinders it from going into mass market. On the contrary, our WINS is very lightweight and low-cost in that it enables the pose estimation for autonomous MAVs upon commodity WiFi.

\section{Feasibility Study}
\label{sec:feasibility}
In this paper, we aim to leverage the drift-free localizability from commodity Wi-Fi to correct the IMU's drift for pose estimation. In principle, with the odometry $\Delta d$ measured by IMU between two positions, the angles with respect to only one AP is enough to locate the MAV via triangulation. 
To study the feasibility of this idea, we conduct an observability analysis and implement a simulator based on the extended Kalman filter (EKF). 

{\bf Observability analysis}. 
Due to the space limitation, we omit the detailed analytical derivation but present the key steps and remarks of the analysis. Please find the full observability analysis in \url{https://github.com/weisgroup/Observability-Analysis}.

Conforming to our system setting, we focus on the case of a single AP and consider the frame of the antenna array attached with a WiFi card is coincided with the IMU frame. The state vector is the $(12\times 1)$ vector. 
\begin{equation}
	\mathbf{x} = \left[\mathbf{q}_w^b; \quad \mathbf{v}_b^w; \quad \mathbf{p}_b^w; \quad \mathbf{c}^w\right],
\end{equation}
where $\mathbf{q}_w^b$ is the unit quaternion that denotes the attitude of the world from $\{w\}$ in the body frame $\{b\}$. The body frame $\{b\}$ is attached to the IMU and $\{w\}$ is a reference frame whose origin coincides with the initial IMU position. $\mathbf{v}_b^w$ and $\mathbf{p}_b^w$ are the 3D position and velocity of $\{b\}$ in $\{w\}$. $\mathbf{c}^w$ denotes the map state, which is the APs' positions.

The observability matrix $\mathbf{O}$ can be defined as
\begin{equation}
	\mathbf{O}(\mathbf{x}^*) = 
	\begin{bmatrix}
		\mathbf{H}_1; \; \mathbf{H}_2\bm{\Phi}_{2, 1}; \; \cdots;  \; \mathbf{H}_k\bm{\Phi}_{k, 1}
	\end{bmatrix},
\end{equation}
where $\bm{\Phi}_{k, 1} = \bm{\Phi}_{k, k-1} \cdots \bm{\Phi}_{2, 1}$ is the state transition matrix from time $1$ to $k$, and $\mathbf{H}_k$ is the measurement Jacobian matrix for the AoA at time $k$. Since all the Jacobians are evaluated at a particular state $\mathbf{x}^* = \left[\mathbf{x}_1^*; \, \cdots; \, \mathbf{x}_k^*\right]$, the observability matrix is a function of $\mathbf{x}^*$. If $\mathbf{O}(\mathbf{x}^*)$ is full column rank, then the model will be fully observable. However, $\mathbf{O}(\mathbf{x}^*)$ is rank deficient. Specifically, when linearizing at the true state, \ie, $\mathbf{x}^* = \mathbf{x}$. The right nullspace $\mathbf{N}$ of the linearized model $\mathbf{O}(\mathbf{x}^*)$ is
\begin{equation}
	\mathbf{N} = 
	\begin{bmatrix}
		\mathbf{0}_3	&	\mathbf{R}(\mathbf{q}_w^{b_1})\mathbf{g}^w	\\
		\mathbf{0}_3	&	-\lfloor\mathbf{v}_{b_1}^w\times\rfloor\mathbf{g}^w	\\
		\mathbf{I}_3	 	&	-\lfloor\mathbf{p}_{b_1}^w\times\rfloor\mathbf{g}^w	\\
		\mathbf{I}_3	 	&	-\lfloor\mathbf{c}^w\times\rfloor\mathbf{g}^w
	\end{bmatrix} = \left[\mathbf{N}_t \quad | \quad \mathbf{N}_r\right],
	\label{eqn:nullspace}
\end{equation}
where $\mathbf{R}(\mathbf{q}_w^{b_1})$ denotes the rotation matrix corresponding to $\mathbf{q}_w^{b_1}$, $\textbf{g}^w = [0, 0, g]^T$ the gravity vector in the earth's inertial frame. The nullspace $\mathbf{N}$ can be easily verified by multiplying each of its blocks with $\mathbf{O}(\mathbf{x}^*)$ as $\mathbf{O}(\mathbf{x}^*)\mathbf{N} = \mathbf{0}$. From Eqn.~\eqref{eqn:nullspace}, we can see the following properties. 
\begin{itemize}
	\item The $(12\times 3)$ block column $\mathbf{N}_t$ corresponds to 3D global translations, \ie, translating both the vehicle and the AP by the same amount.
	\item The $(12\times 1)$ column $\mathbf{N}_r$ corresponds to global rotations of the vehicle and the AP about the gravity.
\end{itemize}

However, in practice, we can only linearize at the estimated state, \ie, $\hat{\mathbf{x}}$. Due to errors in the estimates, the observability matrix gains rank. Specifically, the last two block elements of $\mathbf{O}_k$ are identity matrices, indicating that they are not a function of linearization point. Thus, the nullspace corresponding to translation are preserved, \ie, $\mathbf{O}(\hat{\mathbf{x}})\mathbf{N}_t = \mathbf{0}$. On the contrary, $\mathbf{O}(\hat{\mathbf{x}})\hat{\mathbf{N}}_r \neq \mathbf{0}$ in general since $\hat{\mathbf{N}}_r$ and $\mathbf{O}(\hat{\mathbf{x}})$ depend on the linearization point. In another word, the directions in which the estimator gains information are altered as we use different state estimates to evaluate the system and measurement Jacobians. This makes $\hat{\mathbf{N}}_r$, which corresponds to global rotations, not in the nullspace of $\mathbf{O}(\hat{\mathbf{x}})$. The rank of the observability matrix increases one. 

As a result, the linearized estimator has three unobservable directions, \ie, the ones corresponding to the global translations. In practice, this corresponds to the case when the vehicle moves along the direction of the line connecting itself to the AP. It is conceivable that the AoAs do not change under such motions. Nevertheless, as long as the vehicle does not deliberately keep moving along the unobservable direction, we can update the vehicle's state. Note that when there is no PAoA estimation due to occlusion, there are two cases: 1) the occlusion blocks the direct path PAoA but the WiFi card can still receive packets and estimate the AoA of a reflected path. This is still valid to correct the IMU drift to some extent but the system will be less accurate as shown in Section~\ref{subsec:obstacle}; 2) the occlusion is so strong that the WiFi connection is completely lost. There is no PAoA at all to the vehicle. Then the system cannot be updated so that it becomes completely unobservable.

{\bf EKF Simulation}. Here we run an EKF simulator to verify that the vehicle's state can be updated with AoAs. We deploy $1$ AP and outputs AoAs within $[-90\degree, 90\degree]$, corrupted with Gaussian noise with $1$ degree of standard deviation. The simulated IMU runs at $100$ Hz with $0.01$ $\text{m}/\text{s}^2$ and $0.01$ $\text{rad}/\text{s}$ standard deviation of the additive accelerometer and gyroscope noise. Figure~\ref{fig:verify} shows the estimated trajectory in three cases: IMU only, IMU-AoA fusion with $2$ Hz AoA rate, and IMU-AoA fusion with $20$ Hz AoA rate. We can see that the high-rate AoA succeeded in correcting the IMU's temporal drift. The positioning errors at the end of the trajectory are $0.0293$ m for IMU-AoA ($20$ Hz) fusion, $1.2583$ m for IMU-AoA ($2$ Hz), and $2.6905$ m for IMU only, respectively. The position error of IMU only will increase indefinitely.

In summary, the observability analysis and simulation demonstrate the feasibility and suggests the real-time property of the AoA estimation -- a data rate of tens of hertz. However, EKF fails to work in practice in that 1) it requires good initializations, which are very hard to be obtained by WiFi AoAs due to the lack of metric scale information; 2) EKF early fixes the linearization points may lead to suboptimal results. Therefore, we employ a graph-based SLAM framework and formulate the linear sliding window estimator (Section~\ref{sec:state}).

\section{Positioning AoA Estimation}
\label{sec:wifi}
\subsection{Attitude Estimation}
\label{sec:aoa_disambiguation}
In the context of MAVs, the AoA obtained by the mounted antenna array is a mixed result from MAV translations and rotations. However, only the AoA with respect to translations can be used to localize the MAV. When there is no rotation, the obtained AoA is the PAoA as it only changes with translations. With rotations, if we are aware of the attitude of the MAV $\mathbf{R}_i^0$ at $i\spth$ AoA with respect to the initial pose, where $\mathbf{R}_i^0 \in \mathbb{R}^{3\times 3}$ and $\mathbf{R}_i^0$ orthonormal, the PAoA can be trivially derived. Integrating the angular rate from the onboard IMU obtains the attitude but suffers a temporal drift.

We utilize the insight of multiview reconstruction~\cite{martinec2007robust} in CV to eliminate the rotation drift. We wish to estimate $\mathbf{R}_{k}^{0}$, $k \geq 0$. It subjects to conditions $\mathbf{R}_{0}^{0} = \mathbf{I}_3$, $\mathbf{R}_{j}^{0} = \hat{\mathbf{R}}_{j}^{i}\cdot\mathbf{R}_{i}^{0}$, where $\hat{\mathbf{R}}_{j}^{i}$ denotes a rotation between $i\spth$ and $j\spth$ AoAs. The CV solution \cite{martinec2007robust} finds the essential matrices between the current image and past images to obtain $\hat{\mathbf{R}}_{j}^{i}$. In this context, due to the presence of IMU, $\hat{\mathbf{R}}_{j}^{i}$ can be {\em accurately} obtained by integrating gyroscope measurements when the time window of the two AoAs is small~\cite{6641260}. Specifically, 
\begin{equation}
	 \hat{\mathbf{R}}_{j}^{i}  = \int_{t\in[i, j]}\mathbf{R}_t^{i}\lfloor\hat{\bm{\omega}}_t\times\rfloor\,\mathrm{d}t,
	 \label{eqn:rotation}
\end{equation}
where $\lfloor\hat{\bm{\omega}}_t\times\rfloor$ is the skew-symmetric matrix from the angular velocity $\hat{\bm{\omega}}_t$, $\hat{\bm{\omega}}_t \in \mathbb{R}^{3}$. $\mathbf{R}_t^{i}$ denotes the incremental rotation from $i$ to current time $t$, which is available through short-term integration of gyroscope measurements.

The above system can be solved linearly by relaxing orthonormality constraints of the rotations. Specifically, for a pair of rotation matrices $\mathbf{R}_{i}^{0}$, $\mathbf{R}_{j}^{0}$, and their relative constraint $\hat{\mathbf{R}}_{j}^{i}$, we have 
\begin{equation}
  \begin{bmatrix}
    \mathbf{I}_3, -\hat{\mathbf{R}}_{j}^{i}
  \end{bmatrix}
  \begin{bmatrix}
    \mathbf{r}_{i}^k  \\
    \mathbf{r}_{j}^k
  \end{bmatrix}
  = 0, \quad k = 1, 2, 3,
  \label{eqn:rotate_system}
\end{equation}
\noindent where $\mathbf{r}_{i}^k$ is the $k\spth$ column of $\mathbf{R}_{i}^{0}$. The solution of the relaxed approximate rotation matrices can be found as the last three columns of the right singular matrix of the system \eqref{eqn:rotate_system}, which forms approximate rotation matrices $\bar{\mathbf{R}}_{i}^{0}$. Given the singular value decomposition of the approximate matrix $\bar{\mathbf{R}}_{i}^{0} = \mathbf{U}\mathbf{S}\mathbf{V}^T$, we obtain the true rotation matrices by enforcing unit singular values, $\hat{\mathbf{R}}_{i}^{0} = \mathbf{U}\mathbf{V}^T$. After this point, the rotation estimation of the onboard antenna array is known and drift-free.

\subsection{Real-time Positioning AoA Estimation}
\label{subsec:paoa}
To resolve the multipath components of a signal propagating in indoors, the state-of-the-art technique, SpotFi~\cite{kotaru2015spotfi}, leverages the availability of multiple subcarriers of OFDM modulation scheme to expand the AoA resolution. The problem that prevents it from our pose estimation is that it takes seconds to calculate an AoA. Such a delay causes large errors of pose estimation.

The bottleneck complexity of SpotFi is twofold: 1) The parameter evaluation of all possible AoAs, $\theta \in [-90\degree, 90\degree]$, makes the computational complexity ``anytime'', depending on the resolution step and the range of the evaluation. Reducing the evaluating number of parameters could be effective to improve the computational speed. 2) It accumulates a few packets $(\geq 10)$ to learn the direct path AoA. Using one packet to infer the AoA could be significantly faster.

Our idea is to leverage the motion restriction of MAVs. The intuition is that the next AoA $\theta_{i+1}$ can be bounded by the relative pose from the current AoA $\theta_i$. The relative pose can be obtained from IMU integration due to the fact that the IMU drift is negligible over a small time window~\cite{6641260}. We denote the relative pose as $\mathbf{E}_{i+1}^i = \left[\mathbf{R}_{i+1}^i, \mathbf{p}_{i+1}^i\right]$, where $\mathbf{R}_{i+1}^i \in \mathbb{R}^{3\times 3}$ is the relative rotation, $\mathbf{p}_{i+1}^i \in \mathbb{R}^{3\times 1}$ denotes the relative translation. Given $\theta_i$, instead of evaluating the entire parameter space for $\theta_{i+1}$, searching a small range of parameters based on the prediction of $\theta_{i+1}$ from IMU measurements is more efficient. 

Specifically, the increment from $\theta_i$ to $\theta_{i+1}$ consists of two components: one caused by translation $\theta_t$, the other caused by rotation $\theta_R$. $\check{\theta}_{i+1} = \theta_i + \theta_t + \theta_R$, where $\check{\theta}_{i+1}$ denotes the prediction of $\theta_{i+1}$. $\theta_R$ can be trivially derived from the rotation matrix $\mathbf{R}_{i+1}^i$, which is obtained by Eqn.~\eqref{eqn:rotation}. But $\theta_t$ needs to know the distance $r$ between the antenna array and the AP as the arc length $\|\mathbf{l}_{i+1}^i\| = \theta_t \times r$. Here we assume that the translation is small between two consecutive AoAs. The arc length approximately equals the translation,
\begin{equation}
	\hat{\mathbf{l}}_{i+1}^i \approx \mathbf{p}_{i+1}^i = \mathbf{R}_i^0 \bm{\alpha}_{i+1}^i - \mathbf{R}_i^0 \textbf{g}^{i}\Delta t^2/2,
	\label{eqn:translation}
\end{equation}
\begin{equation}
	\bm{\alpha}_{i+1}^i = \iint_{t\in[i, i+1]}\mathbf{R}_t^{i}\mathbf{a}_t\mathrm{d}t^2,
	\label{eqn:alpha}
\end{equation}
where $\mathbf{a}_t$ is the linear acceleration measured by IMU, $\mathbf{R}_i^0$ can be estimated by the attitude estimation method. $\Delta t$ denotes the time difference between the two AoAs. The gravity $\textbf{g}^{i}$ is initialized as $(0, 0, 9.8)^\intercal$ and updated in the sensor fusion module (see Section~\ref{sec:state}). The unknown initial distance $r$ can be initialized by a reasonable guess. Then $\theta_t = \left\|\hat{\mathbf{l}}_{i+1}^i\right\| / r$. In practice, we found that the solution is very insensitive to this initial value because our proposed sensor fusion model will jointly optimize the positions of the AP and the MAV at runtime, updating $r$ accordingly. At this point, we can predict $\check{\theta}_{i+1}$ from $\theta_i$. Since we have conducted some approximations, we set a fixed smaller range of parameter evaluation $\Delta\theta$ $(= 20\degree)$ to combat the AoA uncertainty of the prediction. 

\begin{algorithm} 
\caption{Real-time PAoA Estimation}
\label{alg:inertialmusic}
\begin{algorithmic}[1]
  \STATE Goal: Output the attitudes of the antenna array and the PAoAs with respect to the initial pose 
  \STATE Initialization: Run SpotFi to obtain the initial AoA $\theta_0$
  \STATE Given $i$\spth AoA $\theta_i$, $(i+1)\spth$ CSI matrix $\mathbf{X}_{i+1}$ measured by the WiFi card, and IMU measurements $\mathbf{a}_t$ and $\bm{\omega}_t$
  
  \STATE $\hat{\mathbf{R}}_{i+1}^i \leftarrow$ Eqn.~\eqref{eqn:rotation} using $\bm{\omega}_t$ 
  \STATE $(\mathbf{R}_i^0, \mathbf{R}_{i+1}^0) \leftarrow \text{attitude\_estimation}(\hat{\mathbf{R}}_{i+1}^i)$
  \STATE $\hat{\mathbf{l}}_{i+1}^i \leftarrow$ Eqn.~\eqref{eqn:translation} and Eqn.~\eqref{eqn:alpha} using $\mathbf{a}_t$ and $\mathbf{R}_i^0$
  \STATE $\hat{\mathbf{R}}_{i+1}^i \rightarrow \theta_R$,  $\hat{\mathbf{l}}_{i+1}^i \rightarrow \theta_t$
  \STATE $\check{\theta}_{i+1} \leftarrow \theta_i + \theta_t + \theta_R$
  \STATE $\mathbf{E}_\text{noise} \leftarrow \text{eigendecomposition}(\mathbf{X_{i+1}}\mathbf{X_{i+1}}^H)$
  \STATE $\mathbf{H} \leftarrow \mathbf{E}_\text{noise}\mathbf{E}_\text{noise}^H$
  \FOR {$\tau = 0:\Delta\tau:\frac{1}{f_\delta}$}  
  	\STATE {\textbf{comments}: The step size $\Delta\tau$ of time-of-flight $\tau$ is in unit of nanosecond}
  	\FOR {$\theta = \check{\theta}_{i+1}-\Delta\theta:\check{\theta}_{i+1}+\Delta\theta$}
	  	\STATE $P_{MU}(\theta, \tau) = \frac{1}{\|\mathbf{s}^H(\theta,\tau)\ast\mathbf{H}\ast\mathbf{s}(\theta,\tau)\|}$
  	\ENDFOR
  \ENDFOR
  \STATE Find the peaks of $P_{MU}$
  \STATE Select the peak with lowest $\tau$ and the corresponding AoA is $\theta_{i+1}$
  \STATE $^a\hat{\mathbf{R}}_{i+1}^i \leftarrow (\theta_i, \theta_{i+1} - \theta_t)$
  \STATE $(\hat{\mathbf{R}}_i^0, \hat{\mathbf{R}}_{i+1}^0) \leftarrow \text{attitude\_estimation}(^a\hat{\mathbf{R}}_{i+1}^i)$ 
  \STATE $\theta_i^0 \leftarrow (\hat{\mathbf{R}}_i^0, \theta_i)$, $\theta_{i+1}^0 \leftarrow (\hat{\mathbf{R}}_{i+1}^0, \theta_{i+1})$
\end{algorithmic}
\end{algorithm}

Our real-time PAoA estimation algorithm is summarized in Algorithm~\ref{alg:inertialmusic}. Initially, we obtain the first AoA $\theta_0$ by running SpotFi which needs to accumulate tens of packets to learn a more accurate result (Line $2$). Then we gather all the available measurements from WiFi and IMU to efficiently estimate the AoAs for the following packets (Line $3$). Based on $\theta_i$, it predicts the next AoA $\check{\theta}_{i+1}$ via IMU measurements (Line $4$ -- $8$). Next, it goes through the steps of the AoA super-resolution algorithm~\cite{kotaru2015spotfi} (Line $9$ -- $18$). The difference is that our algorithm significantly shrinks the parameter searching range based on the predicted AoA $\check{\theta}_{i+1}$ to boost the computational speed (Line $13$). The obtained AoA $\theta_{i+1}$ is drift-free. We can use it to eliminate the attitude drift. From Line $7$, we have obtained the translational angle $\theta_t$. Subtracting it from $\theta_{i+1}$ gives a drift-free rotational angle. It is trivial to convert $\theta_i$ and $\theta_{i+1} - \theta_t$ to a rotation matrix $^a\hat{\mathbf{R}}_{i+1}^i$ (Line $19$). Then Line $20$ applies the attitude estimation approach again to eliminate the drift by the AoA constraint. Finally, we isolate the PAoA with the drift-free attitude (Line $21$).

\section{WiFi-inertial Sensor Fusion}
\label{sec:state}
Our design philosophy is that PAoAs provide additional geometric constraints to the MAV odometry, which can be used to correct the IMU drift by making the pose estimates best match the measured PAoAs in multiple views. The approach takes the IMU measurements, the attitudes and the PAoAs into a common problem where all poses are jointly estimated, thus considering all correlations amongst them. These correlations are key for any high-precision inertial-based autonomous system~\cite{leutenegger2015keyframe}. 

{\bf Notations}: We introduce the world frame $\{w\}$ and the body frame $\{b\}$ in Section~\ref{sec:feasibility}. Here we further denote $(\cdot)^{a_k}$ and $(\cdot)^{b_k}$ as the antenna array frame and the IMU body frame at the time when obtaining $k\spth$ PAoA. Note that the IMU outputs its measurements at a higher rate than the PAoA estimation from WiFi signals, multiple IMU measurements may exist in the interval $[k, k +1]$. $\textbf{g}^{b_k}$ denotes the gravity vector expressed in the IMU body frame when obtaining $k\spth$ PAoA.

\subsection{Sliding Window Formulation}
\begin{figure}[t]
	\centering
	\begin{minipage}[b]{0.27\textwidth}\centering
		\center
		\includegraphics[width=1\textwidth]{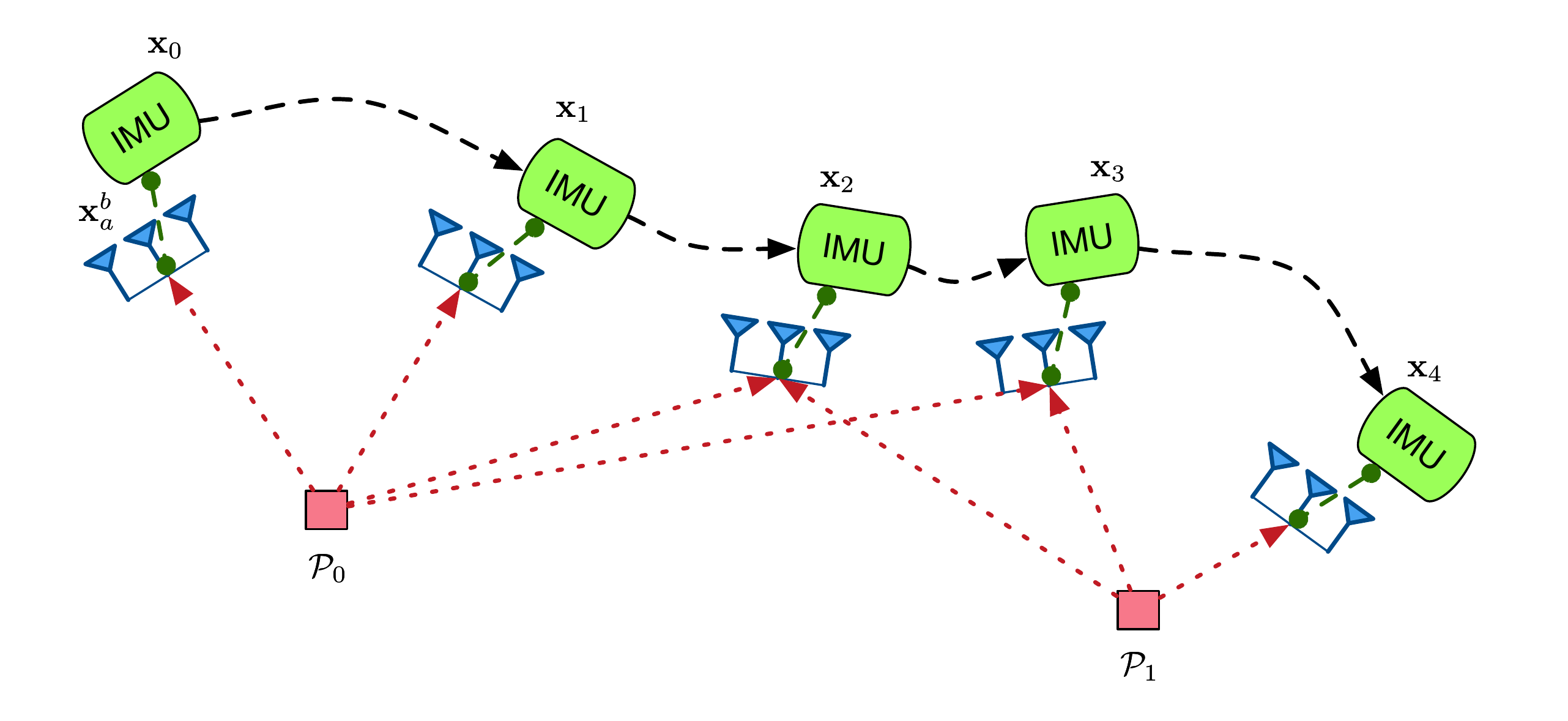}\vspace{-0.3cm}
		\caption{A graph representation of the sliding window.} 
		\label{fig:estimator}
	\end{minipage}
	\hspace{-0.1cm}
	\begin{minipage}[b]{0.21\textwidth}\centering
		\center
		\includegraphics[width=1\textwidth]{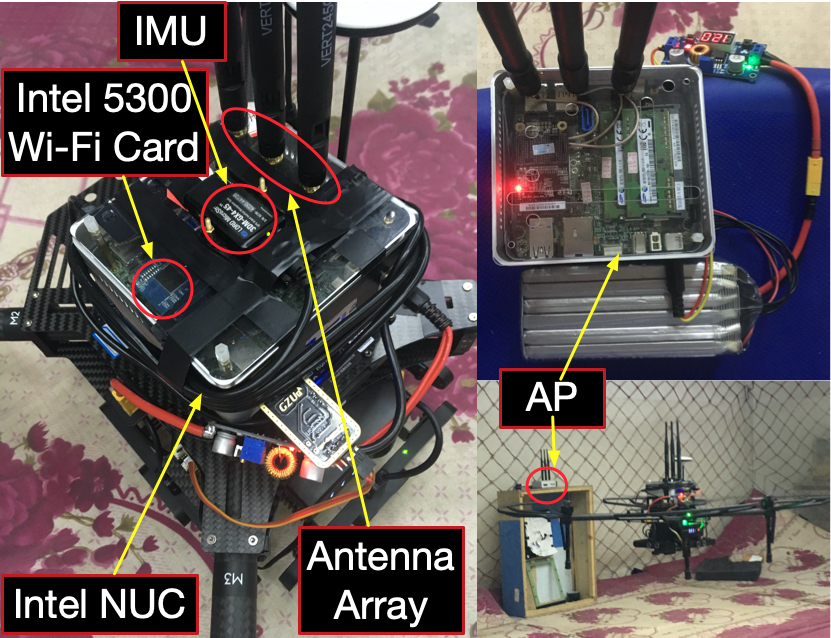}\vspace{-0.3cm}
		\caption{The experimental platform of WINS.} 
		\label{fig:sys}
	\end{minipage}
	\vspace{-0.4cm}
\end{figure}

To make the system run in real-time, we employ the {\em sliding window} formulation~\cite{shen2015tightly, lin2018autonomous} that takes IMU measurements and PAoAs in a fixed time window for pose estimation as shown in Fig.~\ref{fig:estimator}. The window has five frame states $\mathbf{x}_k$ and two APs $\mathcal{P}_l$. Black dotted lines are the pre-integrated IMU measurements and red dotted lines are PAoAs. The unknown array-IMU extrinsic calibration $\mathbf{x}_a^b$ can be manually measured in terms of the relative rotation and translation between them. Although our system works with a single AP, we generalize the formulation to be compatible with multiple APs. The state vector in the sliding window can be defined as, 
\begin{equation}
\begin{aligned}
  \bm{\mathcal{X}} &= [\mathbf{x}_0, \mathbf{x}_1, \dotsc, \mathbf{x}_n, \mathbf{c}_1, \mathbf{c}_2, \dotsc, \mathbf{c}_m]^T          \\
  \mathbf{x}_k     &= [\mathbf{p}_{b_k}^w, \mathbf{v}_{b_k}^w, \textbf{g}^{b_k}]^T, \quad k\in[0, n] \\
  \mathbf{p}_0^0   &= [0, 0, 0]^T,
\end{aligned}
\label{eqn:state}
\end{equation}
\noindent where $\mathbf{x}_k$ is $k\spth$ frame state, which contains position $\mathbf{p}_{b_k}^w$, velocity $\mathbf{v}_{b_k}^w$ in the world frame, and the gravity $\textbf{g}^{b_k}$ in the IMU body frame. 

The world frame is coincident with the real world where the gravity is vertical. It will be set after the gravity vector is solved in the initialization procedure. Similar to the mathematical representation in the PAoA estimation, we also use rotation matrices to represent rotations in this section. $n$ is the number of PAoA frames in the sliding window. State $\mathbf{x}_k$ denotes the state at the time when obtaining $k\spth$ PAoA. $m$ denotes the number of APs in the sliding window and $\mathbf{c}_l$ is $l\spth$ AP's position in the world frame.

Since the IMU outputs its measurements at a higher rate than the PAoA estimation, multiple IMU measurements exist between two consecutive PAoAs. We need to pre-integrate such measurements before the sensor fusion. The IMU preintegration technique was proposed by~\cite{forster2015rss}. Here we give an overview of its usage within WINS.

The global rotation (in the earth's inertial frame) can be determined only with known initial attitude, which is hard to obtain in our system because the metric scale is not directly observable with PAoAs. However, as suggested in~\cite{shen2015tightly}, if the reference frame of the IMU propagation model is attached to the body frame of the first pose that we are trying to initialize, given two time instants that correspond to two consecutive PAoAs, we can write the IMU propagation model for position and velocity in the frame of the first pose as 
\begin{equation}
  \begin{aligned}
    \mathbf{p}_{b_{k+1}}^{b_0}      &= \mathbf{p}_{b_{k}}^{b_0} + \mathbf{R}_{b_{k}}^{b_0}\mathbf{v}_{b_{k}}^{b_k}\Delta t - \mathbf{R}_{b_{k}}^{b_0}\textbf{g}^{b_k}\Delta t^2/2 + \mathbf{R}_{b_{k}}^{b_0}\bm{\alpha}_{b_{k+1}}^{b_k} \\
    \mathbf{v}_{b_{k+1}}^{b_{k+1}}  &= \mathbf{R}_{b_{k}}^{b_{k+1}}\mathbf{v}_{b_{k}}^{b_k} - \mathbf{R}_{b_{k}}^{b_{k+1}}\textbf{g}^{b_k}\Delta t + \mathbf{R}_{b_{k}}^{b_{k+1}}\bm{\beta}_{b_{k+1}}^{b_k}   \\
    \textbf{g}^{b_{k+1}}      &= \mathbf{R}_{b_{k}}^{b_{k+1}}\textbf{g}^{b_{k}},
  \end{aligned}
  \label{eqn:linear_update}
\end{equation}
\noindent where $\Delta t$ is the time difference between the two PAoAs. $\bm{\alpha}_{b_{k+1}}^{b_k}$ and $\mathbf{R}_{b_{k+1}}^{b_k}$ have been defined in Eqn.~\eqref{eqn:alpha} and Eqn.~\eqref{eqn:rotation} except using $(\cdot)^{b}$ notation to clearly indicate the IMU body frame in this section. $\bm{\beta}_{b_{k+1}}^{b_k}$ is defined as
\begin{equation}
	\bm{\beta}_{b_{k+1}}^{b_k}  = \int_{t\in[k, k+1]}\mathbf{R}_t^{b_k}\mathbf{a}_t^b\mathrm{d}t.
	\label{eqn:beta}
\end{equation}
$\bm{\alpha}_{b_{k+1}}^{b_k}$, $\bm{\beta}_{b_{k+1}}^{b_k}$, and $\mathbf{R}_{b_{k+1}}^{b_k}$ in Eqn.~\eqref{eqn:linear_update} can be obtained solely with IMU measurements within $[k, k+1]$. $\mathbf{R}_{b_k}^{b_0}$ is the change in rotation since the first pose (or since the $0\spth$ PAoA), and $\mathbf{R}_{b_{k+1}}^{b_k}$ is the incremental rotation between two consecutive PAoAs. 

At this point, the update equations for all the key quantities will be linear in Eqn.~\eqref{eqn:linear_update} if rotation $\mathbf{R}_{b_k}^{b_0}$ is provided.

\subsection{Linear Sliding Window Estimator}
We change the reference frame of the whole system from the earth's inertial frame to the frame of the first PAoA measurement $b_0$. With this formulation, the dependency on global attitude is removed. The full state vector \eqref{eqn:state} can be rewritten as:
\begin{equation}
  \begin{aligned}
    \bm{\mathcal{X}}           &= [\mathbf{x}_{b_0}^{b_0}, \mathbf{x}_{b_1}^{b_0}, \dotsc, \mathbf{x}_{b_n}^{b_0}, \mathbf{c}_1^{b_0}, \mathbf{c}_2^{b_0}, \dotsc, \mathbf{c}_m^{b_0}]^T   \\
    \mathbf{x}_{b_k}^{b_0}     &= [\mathbf{p}_{b_k}^{b_0}, \mathbf{v}_{b_k}^{b_k}, \textbf{g}^{b_k}]^T, \quad k\in[0, n]                           \\
    \mathbf{p}_{b_0}^{b_0}     &= [0, 0, 0]^T,
  \end{aligned}
\end{equation}
\noindent where $\mathbf{x}_{b_k}^{b_0}$ denotes the frame state when having $k\spth$ PAoA. Again, $n$ is the number of frames in the sliding window, $m$ is the number of APs observed in the sliding window, and $\mathbf{c}_l^{b_0}$ is $l\spth$ AP's position referring to the first pose.

We formulate the linear WINS by gathering all measurements from both the IMU and the antenna array to obtain a maximum likelihood estimation by minimizing the sum of the Mahalanobis norm of all measurement errors 
\begin{equation}
  \begin{aligned}
    \min_{\bm{\mathcal{X}}} & \left\{\left(\mathbf{b}_p - \bm{\Lambda}_p\bm{\mathcal{X}}\right) + \sum_{k\in\mathcal{D}}\left\|\hat{\mathbf{z}}_{b_{k+1}}^{b_k} - \mathbf{H}_{b_{k+1}}^{b_k}\bm{\mathcal{X}}\right\|_{\mathbf{P}_{b_{k+1}}^{b_k}}^2 \right.\\
     &\qquad\left.+\sum_{(l, j)\in\mathcal{A}}\left\|\hat{\mathbf{z}}_{l}^{b_j} - \mathbf{H}_{l}^{b_j}\bm{\mathcal{X}}\right\|_{\mathbf{P}_{l}^{b_j}}^2\right\},
  \end{aligned}
  \label{eqn:cost}
\end{equation}
\noindent where the measurement triplets $\left\{\hat{\mathbf{z}}_{b_{k+1}}^{b_k}, \mathbf{H}_{b_{k+1}}^{b_k}, \mathbf{P}_{b_{k+1}}^{b_k}\right\}$ and $\left\{\hat{\mathbf{z}}_{l}^{b_j}\right.$, $\left.\mathbf{H}_{l}^{b_j}, \mathbf{P}_{l}^{b_j}\right\}$ will be defined in Section~\ref{subsec:imu_mea} and Section~\ref{subsec:aoa_mea} respectively. $\mathcal{D}$ denotes the set of IMU measurements and $\mathcal{A}$ denotes the set of PAoAs within the sliding window. $\left\{\mathbf{b}_p, \bm{\Lambda}_p\right\}$ is the optional prior for the system. To solve the problem~\eqref{eqn:cost}, we next give the measurement model of the IMU and the antenna array.

\subsection{IMU Measurement Model}
\label{subsec:imu_mea}

Given the locally drift-free rotation, we can rewrite \eqref{eqn:linear_update} as a linear function of the state $\bm{\mathcal{X}}$: 
\begin{equation}
  \begin{aligned}
    \hat{\mathbf{z}}_{b_{k+1}}^{b_k} = 
    \begin{bmatrix}
      \hat{\bm{\alpha}}_{b_{k+1}}^{b_k}   \\
      \hat{\bm{\beta}}_{b_{k+1}}^{b_k}    \\
      \hat{\bm{0}}
    \end{bmatrix} 
    &= \begin{bmatrix}
      \mathbf{R}_{b_{0}}^{b_k}\left(\mathbf{p}_{b_{k+1}}^{b_0} - \mathbf{p}_{b_{k}}^{b_0}\right) - \mathbf{v}_{b_{k}}^{b_k}\Delta t + \textbf{g}^{b_k}\frac{\Delta t^2}{2}  \\
      \mathbf{R}_{b_{k+1}}^{b_k}\mathbf{v}_{b_{k+1}}^{b_{k+1}} - \mathbf{v}_{b_{k}}^{b_k} + \textbf{g}^{b_k}\Delta t  \\
      \mathbf{R}_{b_{k+1}}^{b_k}\textbf{g}^{b_{k+1}} - \textbf{g}^{b_k}
    \end{bmatrix} \\
    &= \mathbf{H}_{b_{k+1}}^{b_k}\bm{\mathcal{X}} + \mathbf{n}_{b_{k+1}}^{b_k},
  \end{aligned}
  \label{eqn:imumodel}
\end{equation}
\noindent where $\mathbf{n}_{b_{k+1}}^{b_k}$ denotes the additive measurement noise. We estimate the gravity vector for each pose. The last block line in \eqref{eqn:imumodel} represents the prediction of the gravity vector. All variables except the position component are independent of the accumulated rotation $\mathbf{R}_{b_{0}}^{b_k}$, making them insensitive to rotation error. Typically, we assume the additive noise follows a Gaussian distribution. Then the linear IMU measurement model has the form:
\begin{equation}
  \hat{\mathbf{z}}_{b_{k+1}}^{b_k} \sim \mathcal{N}\left(\mathbf{H}_{b_{k+1}}^{b_k}\bm{\mathcal{X}}, \mathbf{P}_{b_{k+1}}^{b_k}\right),
\end{equation}
where the covariance matrix $\mathbf{P}_{b_{k+1}}^{b_k}$ can be calculated using the pre-integration technique proposed in~\cite{lupton2012visual}.

\subsection{Antenna Array Measurement Model}
\label{subsec:aoa_mea}
The directional vector referred to the $l\spth$ AP observed in the $j\spth$ antenna array frame $\mathbf{d}_{l}^{a_j}$ has the following geometric relationship with respect to the positions of the AP and the MAV,
\begin{equation}
  \lambda_l^{b_j}\left(\mathbf{R}_{a}^{b}\mathbf{d}_{l}^{a_j} + \mathbf{d}_a^b\right) = \lambda_l^{b_j}\mathbf{d}_{l}^{b_j} = \mathbf{R}_{b_0}^{b_j}\left(\mathbf{c}_{l}^{b_0} - \mathbf{p}_{b_j}^{b_0}\right),
  \label{eqn:similar_equation}
\end{equation}
\noindent where $\mathbf{R}_{a}^{b}$ and $\mathbf{d}_a^b$ are the rotation and the translation from the antenna array frame to the IMU body frame, respectively. As we mentioned earlier, $\mathbf{R}_{a}^{b}$ and $\mathbf{d}_a^b$ can be predetermined by manually measuring the relative positions of the antenna array and the IMU. $\lambda_l^{b_j}$ denotes the unknown distance to the AP at this frame, and $\mathbf{c}_{l}^{b_0}$ is the $l\spth$ AP's position referring to the first pose. Thus we use it to transform the PAoA measurement from the antenna frame $\mathbf{d}_{l}^{a_j}$ to the IMU body frame $\mathbf{d}_{l}^{b_j}$. 

Through the geometric relationship between the observed direction and the AP's position, we perform a cross product operation to derive the following expression:
\begin{equation}
  \left(\mathbf{R}_{b_j}^{b_0}\mathbf{d}_{l}^{b_j}\right) \times \left(\mathbf{c}_{l}^{b_0} - \mathbf{p}_{b_j}^{b_0}\right) = \bm{0}.
  \label{eqn:aoa_linear}
\end{equation}
By adding the noise term of PAoA measurements, Eqn.~\eqref{eqn:aoa_linear} can be rewritten as
\begin{equation} 
  \hat{\mathbf{z}}_l^{b_j} = \left\lfloor\mathbf{R}_{b_j}^{b_0}\mathbf{d}_{l}^{b_j}\times\right\rfloor\left(\mathbf{c}_{l}^{b_0} - \mathbf{p}_{b_j}^{b_0}\right) = \mathbf{H}_l^{b_j}\bm{\mathcal{X}} + \mathbf{n}_l^{b_j},
  \label{eqn:wifi_measurement_model}
\end{equation}
\noindent where $\left\lfloor\mathbf{R}_{b_j}^{b_0}\mathbf{d}_{l}^{b_j}\times\right\rfloor$ is the skew-symmetric matrix from $\mathbf{R}_{b_j}^{b_0}\mathbf{d}_{l}^{b_j}$. Again, we assume the noise $\mathbf{n}_l^{b_j}$ follows a Gaussian distribution. Thus, the PAoA measurement model has the form
\begin{equation} 
\hat{\mathbf{z}}_l^{b_j} \sim \mathcal{N}\left(\mathbf{H}_l^{b_j}\bm{\mathcal{X}}, \mathbf{P}_l^{b_j}\right),
\end{equation}
\noindent where $\mathbf{P}_l^{b_j} = {\lambda_{l}^{b_j}}^2 \bar{\mathbf{P}}_{l}^{b_j}$, $\lambda_{l}^{b_j}$ is the distance to the $l\spth$ AP at the $b_j$ frame and $\bar{\mathbf{P}}_{l}^{b_j}$ denotes the PAoA observation noise. Note that $\lambda_{l}^{b_j}$ is unknown initially, and we initialize it with an identical value. In practice, we found that the solution is very insensitive to the initial value of $\lambda_{l}^{b_j}$.

\section{System Evaluation}
\label{sec:evaluation}
\subsection{Implementation and Experimental Setup}
We implement WINS on the Intel NUC with a $1.3$ GHz Core i5 processor with $4$ cores, $8$ GB of RAM and a $120$ GB SSD, running Ubuntu Linux equipped with an Intel 5300 NIC and a LORD MicroStrain 3DM-GX4-25 IMU. We build on the Linux 802.11 CSI tool~\cite{halperin2011tool} to obtain the wireless channel information for each packet. The CSI is available for any WiFi packet no matter the packet is for transmitting control commands or data. The CSI tool extracts the CSI from these transmissions without harming data integrity. Therefore, our system has no interruption on conventional MAV command/data transmissions. The experimental platform is shown in Fig.~\ref{fig:sys}.

To improve the accuracy of PAoA estimation, we use a Vector Network Analyzer to conduct a one-time calibration of transceiver responses of WiFi chips~\cite{kotaru2017position}. We utilize Robot Operating System (ROS) as the communication middleware. On the NUC, the $200$ Hz IMU measurements as well as $50$ Hz PAoAs (Section~\ref{sec:wifi}) are fused in the linear WiFi-inertial pose estimator (Section~\ref{sec:state}) to obtain the MAV poses. In all experiments, we only use one AP. The experiments are conducted in the $12\times 8$ square meters of a MAV test site in our laboratory.

\subsection{Micro-benchmark Evaluation}
\subsubsection{Experiments without rotations}
\label{subsubsec:sae}

We set the attitude of the antenna array and fix the array attitude from $-90$ to $90$ degree in a step size of $10$ degrees. At each step, the AP keeps sending packets in $30$ Hz to the NUC for exactly $30$ seconds. We compare the accuracy of the proposed PAoA estimation algorithm with the state-of-the-art, SpotFi~\cite{kotaru2015spotfi}. Since SpotFi needs to accumulate a set of packets to identify the direct path AoA, in particular, we collect $10$ consecutive packets for that and group the same number of packets for the PAoA estimation. As our algorithm is able to obtain the AoA for each packet, we analyze the result in two ways. One inputs the average result of the AoAs from these $10$ packets for a fairly comparison with SpotFi. The other reports the AoA error from each individual packet. 

\begin{figure}[t!]
    \centering
    \shortstack{
            \includegraphics[width=0.22\textwidth]{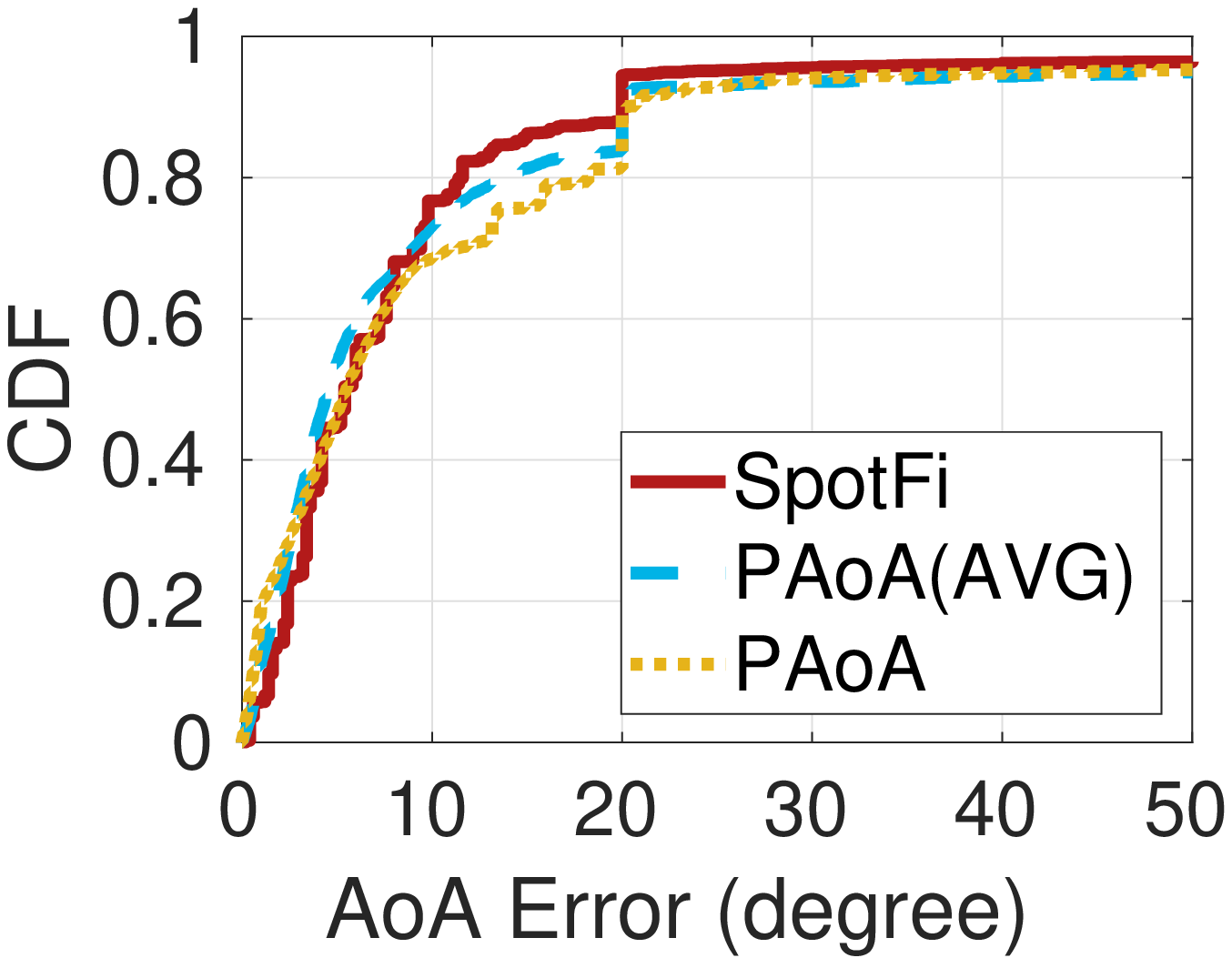}\\
            {\footnotesize (a) CDF of AoA error}
    }\quad
    \shortstack{
            \includegraphics[width=0.22\textwidth]{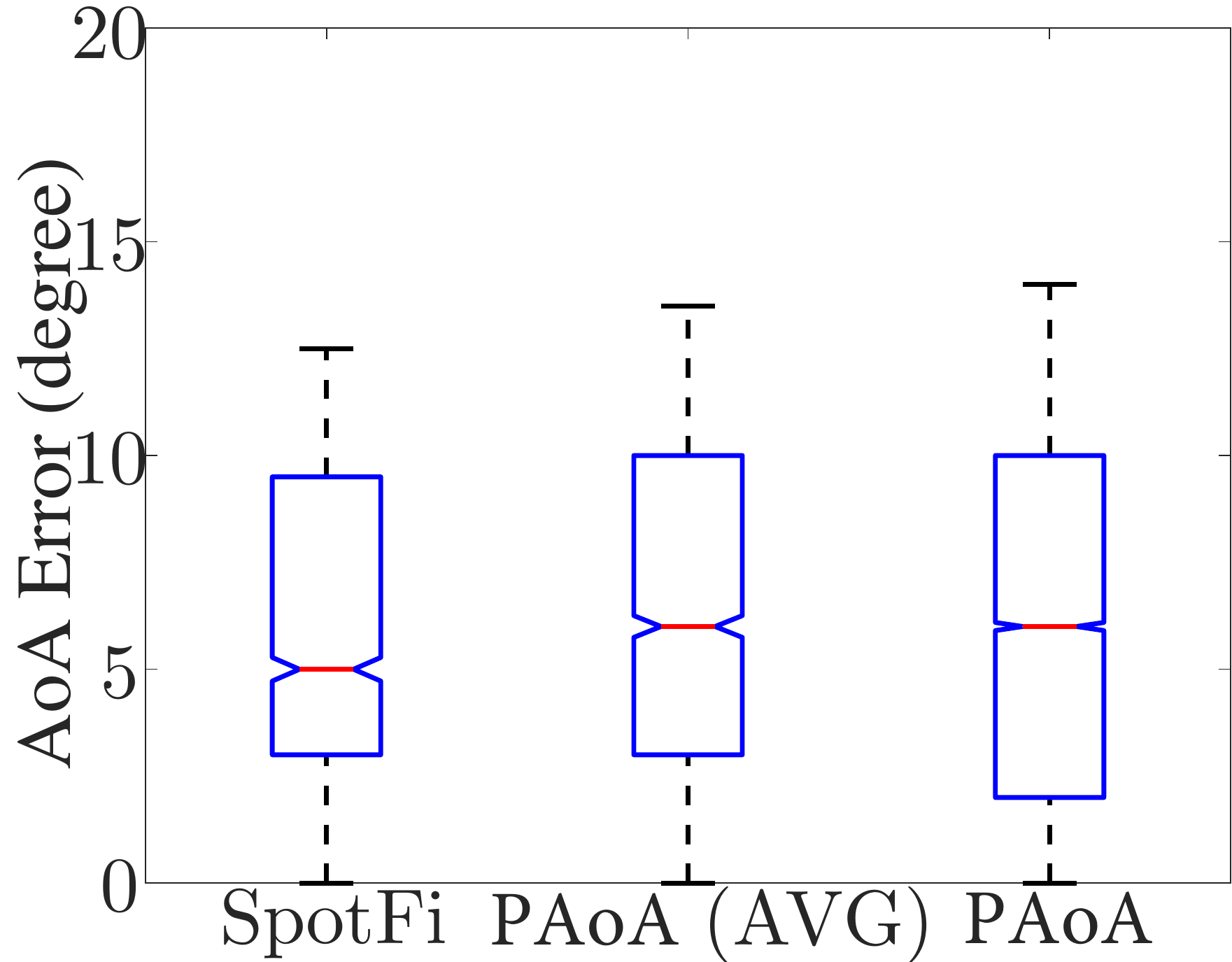}\\
            {\footnotesize (b) Box plot of AoA error}
    }
    \caption{Performance analysis of AoA estimation when the antenna array has no rotation. \label{fig:cdf} } 
    \vspace{-4mm}
\end{figure}

Fig.~\ref{fig:cdf}~(a) plots the CDFs for the error of AoA estimation for SpotFi, PAoA (AVG), and PAoA. PAoA (AVG) denotes the way that AoA estimates of a set of packets are averaged out to be the final result. As expected, incorporating the inertial measurements makes the accuracy of PAoA comparable to SpotFi. 
Fig.~\ref{fig:cdf}~(b) is the box plot of these three approaches. On each box, the central mark indicates the median, and the bottom and top edges of the box indicate the $25\spth$ and $75\spth$ percentiles respectively. This also demonstrates the high accuracy of PAoA.

\subsubsection{Experiments with rotations}

\begin{figure}[t!]
    \centering
    \shortstack{
            \includegraphics[width=0.12\textwidth]{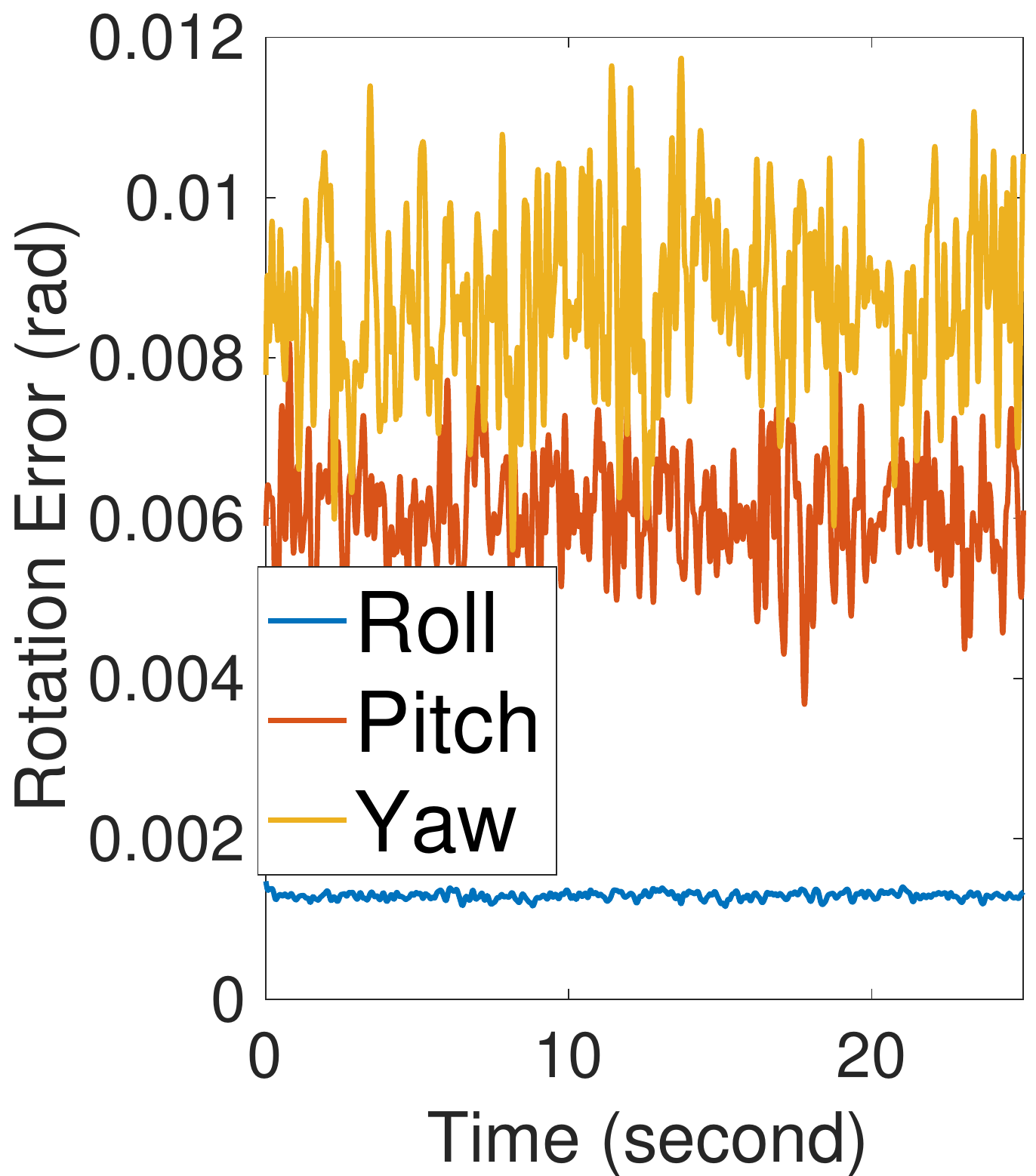}\\
            {\footnotesize (a) Attitude error }
    }\quad
    \shortstack{
            \includegraphics[width=0.3\textwidth]{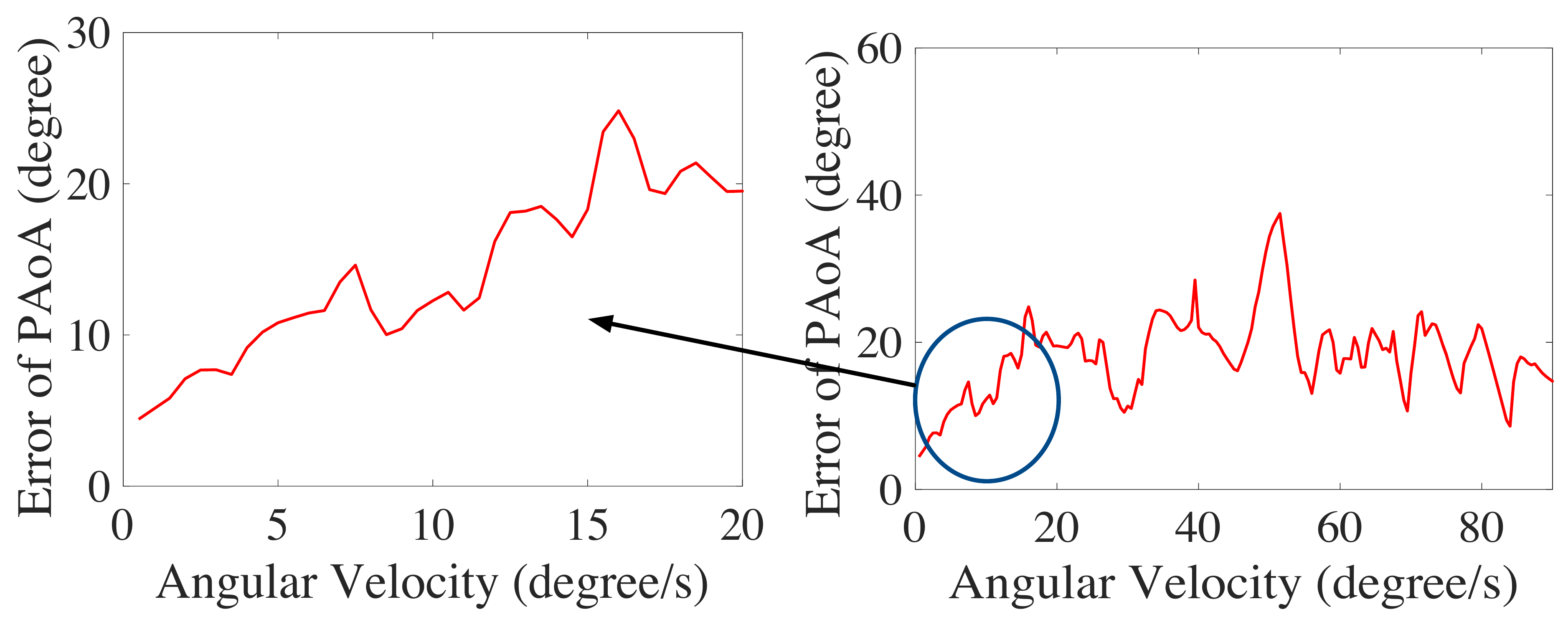}\\
            {\footnotesize (b) Rotary PAoA}
    }
    \caption{Performance analysis of attitude and PAoA estimation with rotary antenna array. \label{fig:mobile} } 
    \vspace{-4mm}
\end{figure}

Due to the fast dynamics of MAVs, it is critical to examine the performance of PAoA estimation when the antenna array rotates. In this experiment, we first examine the performance of attitude estimation. We conduct the experiment by putting the MAV on a rotating table driven by a stepper motor. The ground truth is provided by the controller of the table. After a trivial calibration to setup the initial attitude, the statistical error report of attitude estimation along $25$ seconds over $50$ repeated experiments is shown in Fig.~\ref{fig:mobile}~(a). The yaw angle suffers the drift. The maximum drift within the rotary period is only $0.016$ rad (= $0.92\degree$), demonstrating the drift-free property.

Given the attitude, we obtain the PAoA for each incoming packet. We analyze the collected data and draw the relationship between the angular velocity and the error of PAoA estimation as shown in Fig.~\ref{fig:mobile}~(b). When the angular velocity increases from $0$, the PAoA error exhibits a linear increase. This is because the channel response (CSI) changes fast when the array keeps rotating. Fortunately, the PAoA error does not increase indefinitely along the angular velocity. We can see that the error fluctuates between $10$ and $20$ degrees when the angular velocity goes beyond even $80$ degrees per second. This demonstrates the reliability of the PAoA estimation algorithm that the PAoA error can be bounded with fast dynamics.

\subsubsection{System latency}

\begin{table}[t!]
\footnotesize
  \centering
  \caption{PAoA computation time.}
  \begin{tabular}{ | p{1.5cm} | p{1.4cm} | p{1.8cm} | p{0.8cm} |}
    \hline
    Algorithms & Accumulated Packets & Mean Computation Time (ms) & STD (ms) \\ \hline
    SpotFi       			& 10 & 508.5 & 1206.9 \\ \hline
    PAoA (AVG)			& 10 & 196.3 & 23.7   \\ \hline
    PAoA	 				& 1  & 20.3  & 60.3   \\ \hline
  \end{tabular}
  \label{tab:time}
  \vspace{-4mm}
\end{table}

For the PAoA estimation, we introduce inertial sensing to reduce the computational cost of AoA estimation. We summarize the computation time of PAoA and the state-of-art AoA estimation technique, SpotFi~\cite{kotaru2015spotfi}, in Table~\ref{tab:time}. PAoA (AVG) denotes the same algorithm setting as before. From the rows of SpotFi and PAoA (AVG), we can see that accumulating a set of packets to find a PAoA takes hundreds of milliseconds. This will output a PAoA over a second when employing multiple APs. But when the PAoA algorithm estimates a PAoA for each packet, the computation time is reduced by an order of magnitudes. The last row of the table shows that the PAoA estimation algorithm boosts the computational speed about $25\times$ faster than SpotFi. 

For the sensor fusion, our sliding window scheme intends to ensure the real-time processing. In principle, the more frames involved in the window the more accurate result obtained. But this also increases the computation time because a larger state vector and the corresponding measurements are involved. Therefore, we need to balance the accuracy and the latency of our system. We tune the number of frames in the sliding window from $10$ to $50$ for testing. Due to the space limitation, we omit the detailed result but summarize that the increase of the accuracy becomes marginal but the latency goes up to hundreds of milliseconds when incorporating more than $30$ frames. Therefore, we set $30$ frames for our experiments and the average latency is $37.19$ ms for each update, ensuring the real-time processing.

\subsection{Pose Estimation}
We run WINS on the Intel NUC attached on the DJI Matrice100. An AP is placed at a height of $1.5$ m. The experiment is to control the MAV flying in a circular pattern in the test site of our laboratory at about $1.5$ m height. We jointly estimate $30$ frames in the sliding window. We configure a wireless network by ROS to build a connection between a laptop and the onboard NUC so that we can remotely control the airborne computer. This enables us to run and recompile the airborne code while the quadrotor is flying. In the experiments, we use the measurements from WINS for the MAV control. The actual flight trajectory (ground truth) is captured by the OptiTrack. Therefore, the following experiments show that WINS is applicable to the MAV real-time control task. 

\begin{figure}[t!]
    \centering
    \shortstack{
            \includegraphics[width=0.15\textwidth]{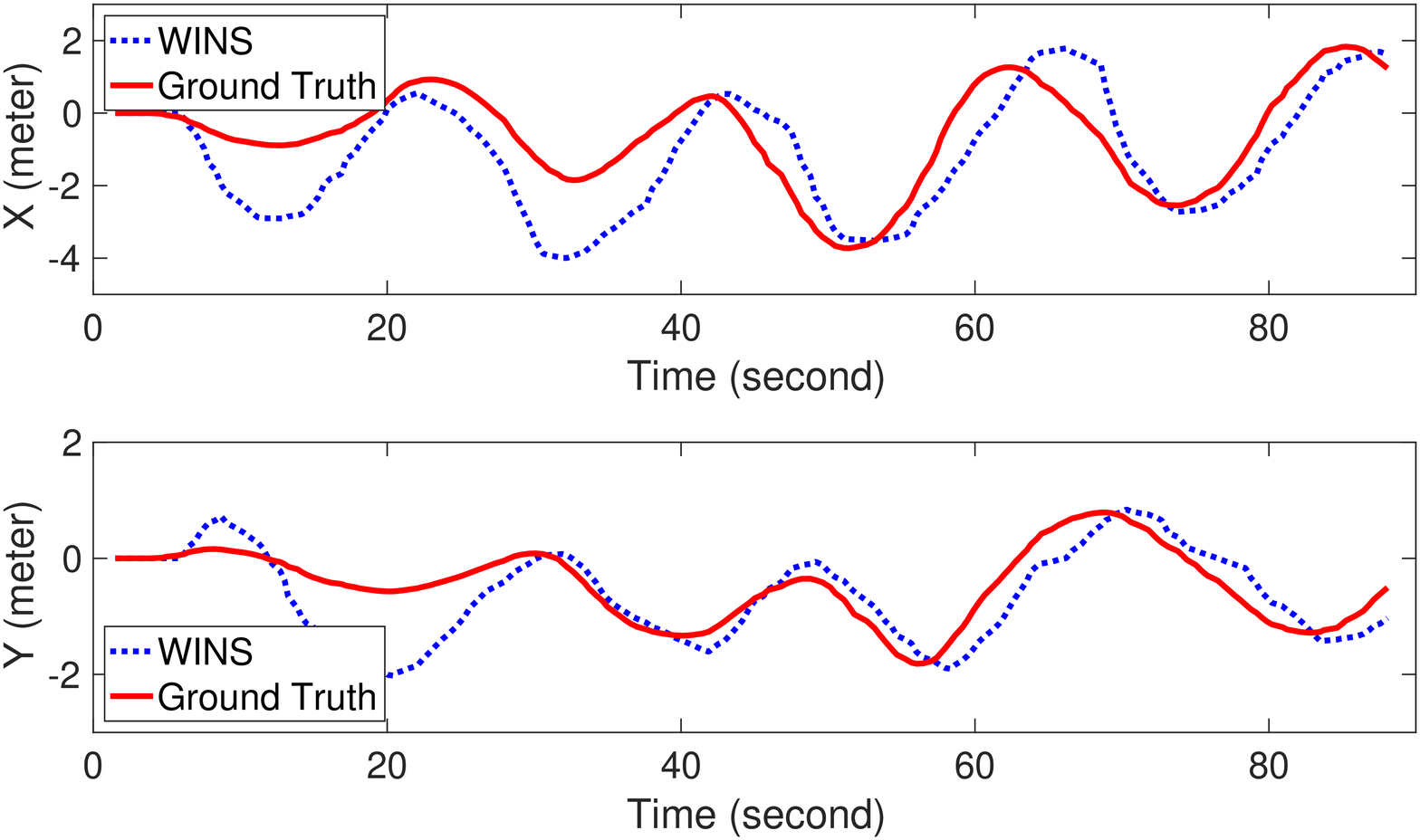}\\
            {\footnotesize (a) MAV position}
    }\quad
    \shortstack{
            \includegraphics[width=0.15\textwidth]{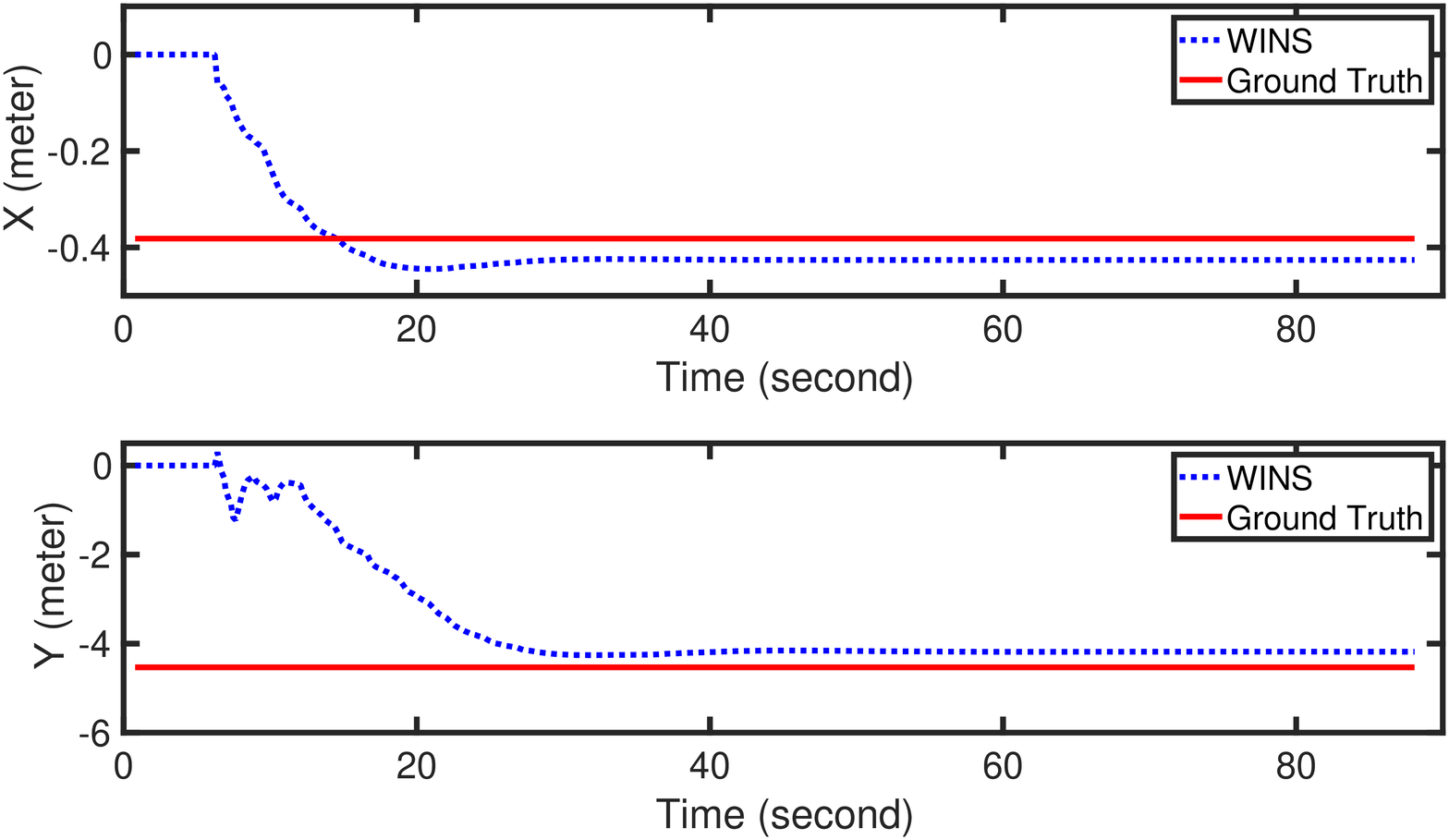}\\
            {\footnotesize (b) AP position}
    }\quad
    \shortstack{
            \includegraphics[width=0.14\textwidth]{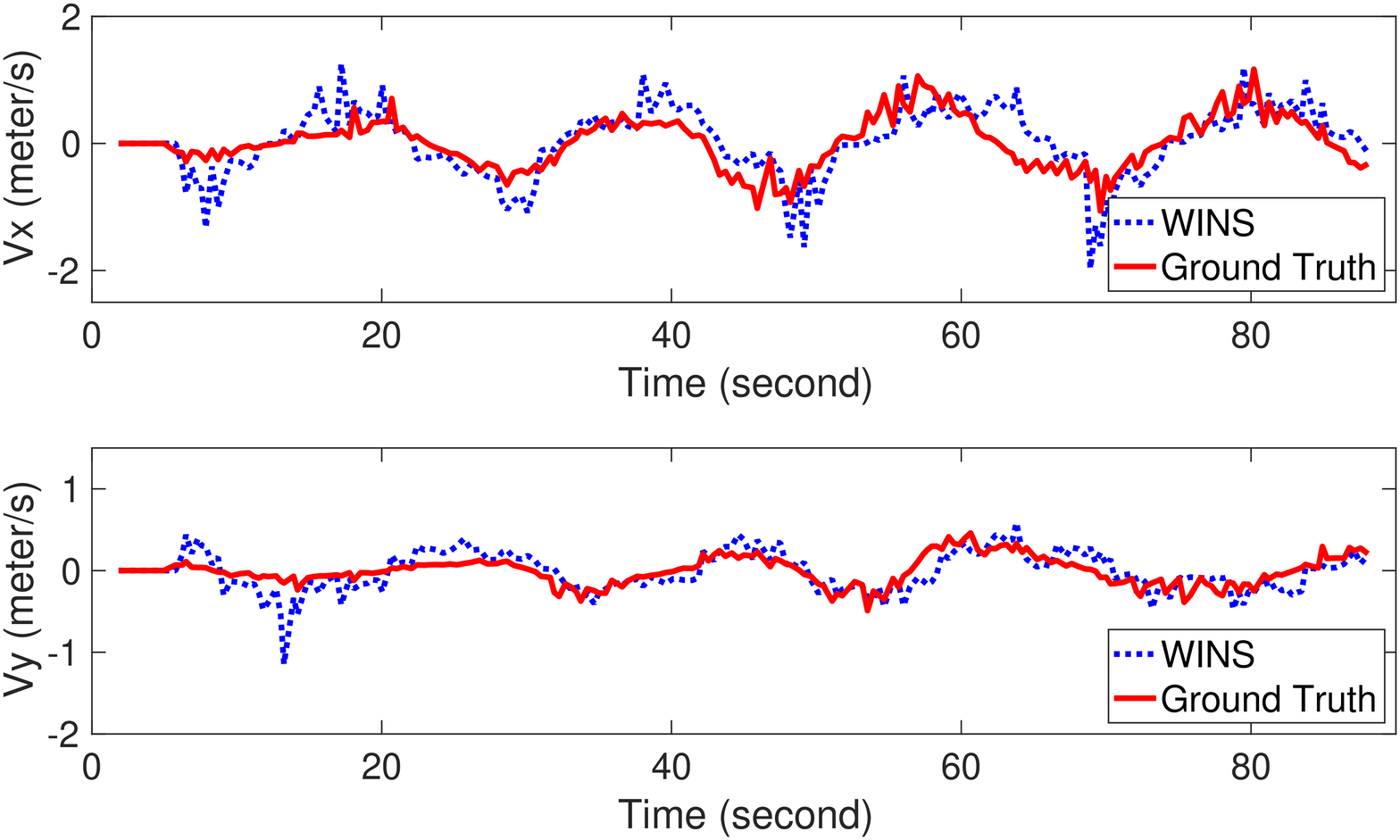}\\
            {\footnotesize (c) Linear velocity}
    }
    \caption{The experimental results of which a MAV flies in a circular pattern.}
    \label{fig:experiment_result}
    \vspace{-4mm}
\end{figure}

Fig.~\ref{fig:experiment_result} shows that WINS is capable of estimating MAV poses with only $1$ AP. Please note that only the azimuth angle is available from our {\em linear} antenna array. Therefore, WINS currently estimates 5-DoF poses, being unable to determine the height of the MAV. The positions of the MAV and the AP are initialized to be $\mathbf{0}$. WINS takes about $30$ seconds to converge, and the trajectory is close to the ground truth eventually. This again demonstrates the online initialization capability of WINS that the initial position can be recovered even randomly choosing an initial value, \eg, $\mathbf{0}$, to bootstrap the system. The mean positioning error of the converged period (after $30$ seconds) is $61.7$ cm, and the standard deviation of $x$-$y$ coordinate is $51.96$ cm and $18.2$ cm, respectively. Fig.~\ref{fig:experiment_result}~(b) shows the performance of AP positioning. The mean position error over the entire flight is $53.4$ cm. From Fig.~\ref{fig:experiment_result}~(c) we can see that the maximum linear velocity reaches $1.2663$ m/s in this experiment. 

\subsection{Obstacle Effects}
\label{subsec:obstacle}

\begin{figure}
	\centering
	\begin{minipage}[b]{0.45\textwidth}\centering
		\center
		\shortstack{
            \includegraphics[width=0.45\textwidth]{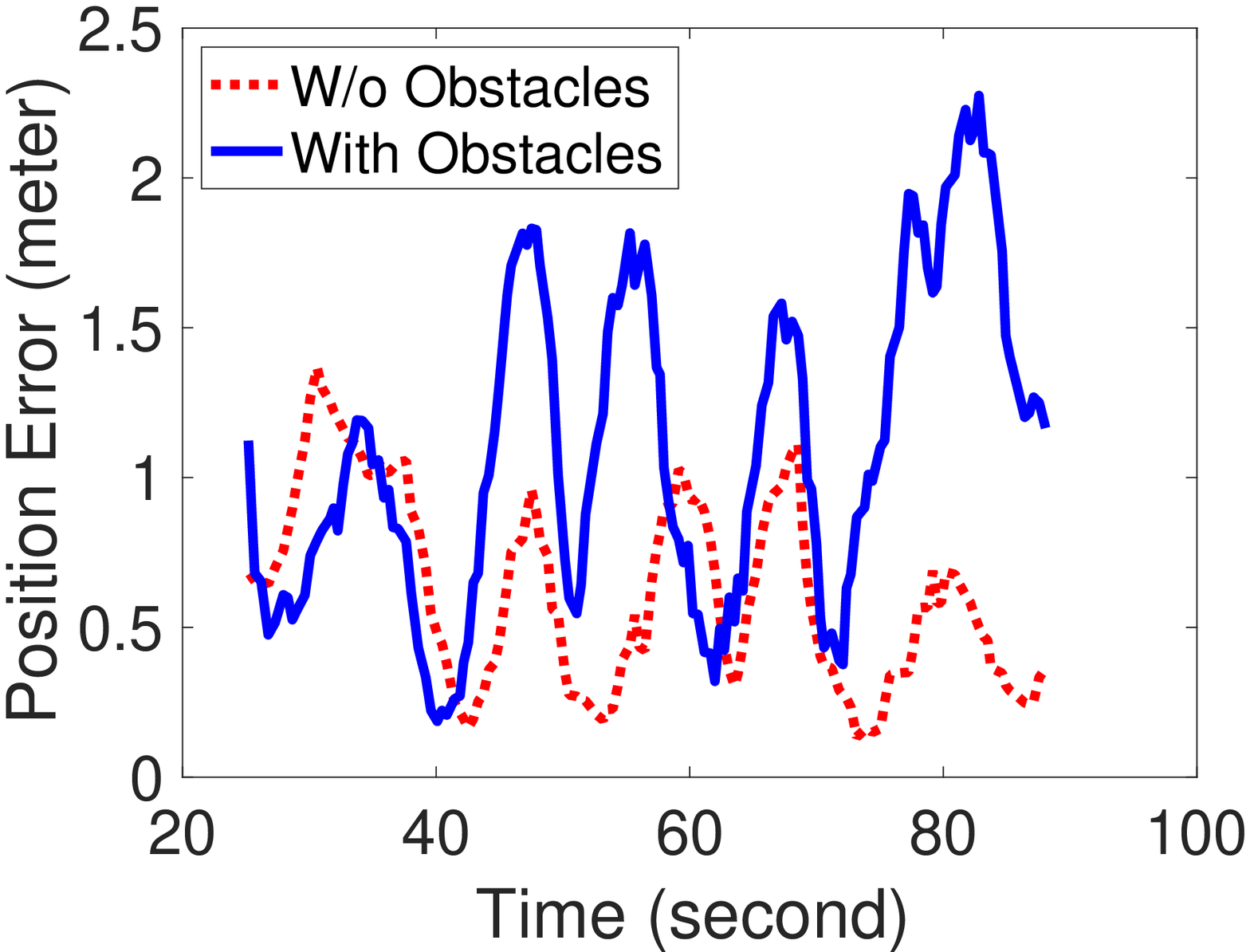}\\
            {\footnotesize (a) MAV position}
    	}\quad
    	\shortstack{
            \includegraphics[width=0.45\textwidth]{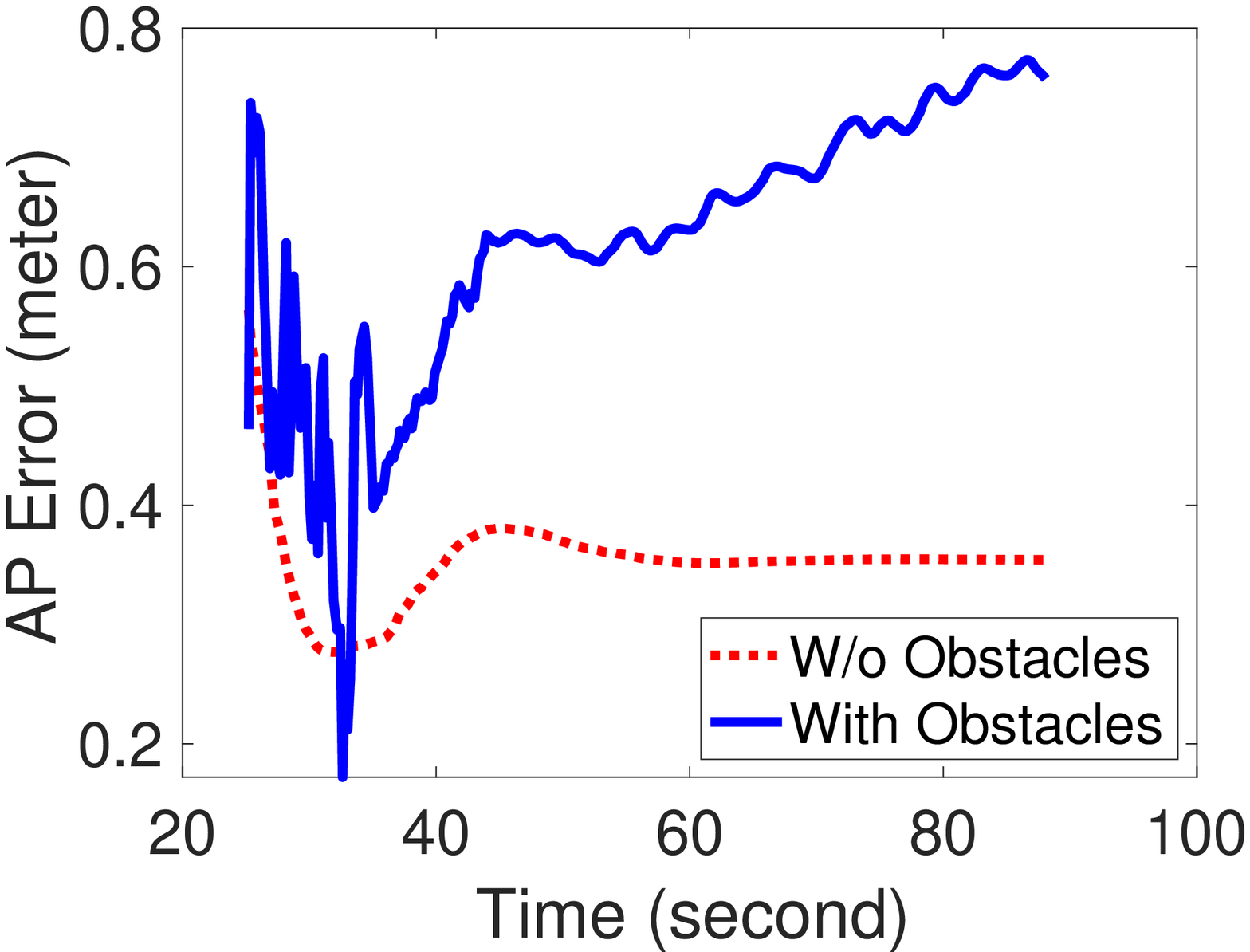}\\
            {\footnotesize (b) AP position}
    	}\quad
    	\caption{The performance effect in the presence of obstacles.} 
    	\label{fig:nlos}
	\end{minipage}
	\vspace{-4mm}
\end{figure}

Theoretically, the signal penetrating obstacles will be received at a lower power level due to the greater signal attenuation. According to the signal attenuation model~\cite{faria2005modeling}, the received power $Pr(\alpha, d)$ (in dBm) at a distance $d$ (in meters) from the transmitter is $Pr(\alpha, d) = Pr_0 - 10\alpha\log(d) + X_\sigma$. $\alpha$ is known as the path loss exponent. It is significantly larger when signals go through walls or other obstacles. The received low-power signal makes the CSI reported by the WiFi NIC less reliable so that the accuracy of AoA estimation is degenerated.

To demonstrate this, we additionally conduct experiments by adding a large board in front of the AP to block the AP and the MAV. As expected, the accuracy is worse with obstacles as shown in Fig.~\ref{fig:nlos}. We plot the result after $25$ seconds when the online initialization has been completed. The accuracy of MAV position is $61.7$ cm and $99.8$ cm without and with obstacles, respectively. The accuracy of AP position is $35.16$ cm and $70.56$ cm without and with obstacles, respectively. The accuracy in the presence of obstacles is still decimeter-level and it is effective to support the navigation.

\subsection{Practical Concerns}
WINS is a prototype of WiFi-inertial pose estimator for MAVs. Here we discuss the practical issues of our prototype in three-fold: the decimeter-level accuracy, the miniature implementation of WINS, and the support of multiple MAVs.

First of all, the decimeter-level accuracy of WINS with a single AP in indoors can be used for many applications like photography and inspection in spacious indoor venues where very few AP are available or deployed, \eg, warehouse and oil tanks~\cite{walmart, mikhail2019flyzone}. Our system can work with only a single AP that can be integrated into the remote controller of the vehicle. For the case of working in confined areas, our proposed approach is practical by trivially improving the accuracy with multiple available APs. In practice, there are often more than one APs available in indoors. It is conceivable that more APs can definitely improve the accuracy. To demonstrate that, we additionally conduct experiments to investigate the accuracy in terms of different available APs. Due to the limited space, we omit the figure and the result shows that the accuracy can reduce to $25.3$ cm with three APs. 

Moreover, WINS also can be easily implemented on a miniature platform with some engineering efforts. The required sensors in our system, which include an IMU and a wireless card, already have miniature designs and they are available in miniature platforms like Crazyflie~\cite{crazyflie}. The computation power of Crazyflie is enough as it can run a deep neural network for visual processing as demonstrated by~\cite{palossi201964mw}. Our WINS is a linear estimator that can be computed quite efficient. Specifically, we use the standard Cholesky decomposition implemented by Eigen to solve the linear system. The time complexity is $O(N^3)$ in theory, where $N$ denotes the number of equations. In practice, the multithreaded routines make the computation time be approximate $N^2$ growth. We have succeeded in running WINS in Raspberry Pi 3, which is a single board computer. The hurdle of the realization in a miniature platform is the small antenna design that should be integrated into the PCB of the platform. This is out of the scope of this paper. 

Finally, WINS supports pose estimation for multiple MAVs with a single AP. WINS is running onboard, meaning that each vehicle works independently. Each vehicle connects to the AP and extracts the CSI of received packets for running WINS, estimating the pose on its own sensor readings. In terms of distinguishing different vehicles, WINS works similar to the case of smartphones that multiple devices can connect to the same AP to access the Internet. The 802.11 MAC frame has address fields to distinguish multiple vehicles through the MAC addresses of their wireless cards.

\subsection{Limitations}

WINS has demonstrated the ability of pose estimation for MAVs using WiFi in indoors. However, our current prototype still has few limitations:

\begin{itemize}
	\item {\em AoA estimation range}. Since the system uses a linear antenna array, the AoA estimation is within $[-90\degree, 90\degree]$. Therefore, the maximum allowable maneuver angles for the MAV is $180\degree$. This can be easily resolved by changing the shape of the antenna array to be a circular array, expanding the AoA estimation range to be $[-180\degree, 180\degree]$. This will slightly change the steering vector, which is very incremental to our work.
	\item {\em Accuracy}. WINS achieves decimeter-level positioning accuracy. It is not designed to defeat the accuracies of CV/laser based approaches~\cite{shen2015tightly, lin2018autonomous, dube2017online}, which can achieve centimeter-level when environments are suitable, \eg, well-lighted, texture-rich, clear in line of sight (LOS). But our design complements them to support navigation in a lightweight and low-cost manner and we are highly resilient to LOS limitations so as to work in vision/laser-crippled scenarios. To improve the accuracy, we can incorporate more APs to jointly optimize the result or incur wide bandwidth signals to enable accurate ranging. We left this to our future work.
	\item {\em Obstacle avoidance}. To fully support the navigation, the indoor environment mapping and obstacle avoidance is crucial besides pose estimation. A promising RF-based approach is to sweep WiFi channels and stitch signals over all available channels to be a wide bandwidth signal, mimicking a mmWave radar to imaging indoor venues and aid in obstacle avoidance. This also leverages ubiquitous WiFi and complements existing line-of-sight sensing modalities, \eg, visual and laser sensing. Evaluating the imaging performance of this approach is out of scope.
\end{itemize}

\section{Conclusion}
\label{sec:conclusion}
We presented WINS, a WiFi-inertial pose estimator with commercial WiFi support. WINS consisted of a real-time PAoA estimation algorithm and a tightly-coupled sliding window sensor fusion model. WINS was lightweight, real-time and instantly deployable. It leveraged WiFi as a new sensing modality to correct the IMU drift to achieve pose estimation. We implemented WINS on the DJI Matrice100 platform equipped with an IMU and a three-antenna linear array. The experiments demonstrated its ability of accurate pose estimation in indoors. We believe that extending our system to fuse more sensing modalities for robust pose estimation in cluttered environments is an important task for future work.

\bibliographystyle{IEEEtranTIE}
\bibliography{IEEEabrv, BIBTIE}

\begin{thebibliography}{10}
\providecommand{\url}[1]{#1}
\csname url@samestyle\endcsname
\providecommand{\newblock}{\relax}
\providecommand{\bibinfo}[2]{#2}
\providecommand{\BIBentrySTDinterwordspacing}{\spaceskip=0pt\relax}
\providecommand{\BIBentryALTinterwordstretchfactor}{4}
\providecommand{\BIBentryALTinterwordspacing}{\spaceskip=\fontdimen2\font plus
\BIBentryALTinterwordstretchfactor\fontdimen3\font minus
  \fontdimen4\font\relax}
\providecommand{\BIBforeignlanguage}[2]{{%
\expandafter\ifx\csname l@#1\endcsname\relax
\typeout{** WARNING: IEEEtran.bst: No hyphenation pattern has been}%
\typeout{** loaded for the language `#1'. Using the pattern for}%
\typeout{** the default language instead.}%
\else
\language=\csname l@#1\endcsname
\fi
#2}}
\providecommand{\BIBdecl}{\relax}
\BIBdecl

\bibitem{walmart}
``Drones will soon take over inventory checks,''
  \url{https://www.inc.com/business-insider/walmart-develops-drones-for-inventory-checks.html},
  {O}nline; available on 31-December-2019.

\bibitem{lin2019kalman}
Z.~Lin, H.~H. Liu, and M.~Wotton, ``Kalman filter-based large-scale wildfire
  monitoring with a system of uavs,'' \emph{IEEE Trans. Ind. Electron.},
  vol.~66, no.~1, pp. 606--615, 2019.

\bibitem{xiao2019sensor}
X.~Xiao, W.~Wang, T.~Chen, Y.~Cao, T.~Jiang, and Q.~Zhang, ``Sensor-augmented
  neural adaptive bitrate video streaming on uavs,'' \emph{IEEE Trans.
  Multimedia}, 2019.

\bibitem{he2019state}
S.~He, W.~Wang, H.~Yang, Y.~Cao, T.~Jiang, and Q.~Zhang, ``State-aware rate
  adaptation for uavs by incorporating on-board sensors,'' \emph{IEEE Trans.
  Veh. Technol.}, 2019.

\bibitem{islam2018observer}
S.~Islam, P.~X. Liu, and A.~El~Saddik, ``Observer-based adaptive output
  feedback control for miniature aerial vehicle,'' \emph{IEEE Trans. Ind.
  Electron.}, vol.~65, no.~1, pp. 470--477, 2018.

\bibitem{lin2018autonomous}
Y.~Lin, F.~Gao, T.~Qin, W.~Gao, T.~Liu, W.~Wu, Z.~Yang, and S.~Shen,
  ``Autonomous aerial navigation using monocular visual-inertial fusion,''
  \emph{J. Field Robot.}, vol.~35, no.~1, pp. 23--51, 2018.

\bibitem{fu2017robust}
Q.~Fu, Q.~Quan, and K.-Y. Cai, ``Robust pose estimation for multirotor {UAVs}
  using off-board monocular vision,'' \emph{IEEE Trans. Ind. Electron.},
  vol.~64, no.~10, pp. 7942--7951, 2017.

\bibitem{tang2019vision}
Y.~Tang, Y.~Hu, J.~Cui, F.~Liao, M.~Lao, F.~Lin, and R.~S. Teo, ``Vision-aided
  multi-uav autonomous flocking in {GPS}-denied environment,'' \emph{IEEE
  Trans. Ind. Electron.}, vol.~66, no.~1, pp. 616--626, 2019.

\bibitem{fichtinger2015assessing}
J.~Fichtinger, J.~M. Ries, E.~H. Grosse, and P.~Baker, ``Assessing the
  environmental impact of integrated inventory and warehouse management,''
  \emph{Int. J. Prod. Econ.}, vol. 170, pp. 717--729, 2015.

\bibitem{kotaru2015spotfi}
M.~Kotaru, K.~Joshi, D.~Bharadia, and S.~Katti, ``Spotfi: Decimeter level
  localization using wifi,'' in \emph{Proc.~ACM SIGCOMM}, 2015.

\bibitem{luo2019dynamic}
R.~C. Luo and T.~J. Hsiao, ``Dynamic wireless indoor localization incorporating
  with an autonomous mobile robot based on an adaptive signal model
  fingerprinting approach,'' \emph{IEEE Trans. Ind. Electron.}, vol.~66, no.~3,
  pp. 1940--1951, 2019.

\bibitem{mueller2015fusing}
M.~W. Mueller, M.~Hamer, and R.~D'Andrea, ``Fusing ultra-wideband range
  measurements with accelerometers and rate gyroscopes for quadrocopter state
  estimation,'' in \emph{Proc.~IEEE ICRA}, 2015.

\bibitem{saab2016novel}
S.~S. Saab and H.~Msheik, ``Novel {RFID}-based pose estimation using single
  stationary antenna,'' \emph{IEEE Trans. Ind. Electron.}, vol.~63, no.~3, pp.
  1842--1852, 2016.

\bibitem{james2018liftoff}
J.~James, V.~Iyer, Y.~Chukewad, S.~Gollakota, and S.~B. Fuller, ``Liftoff of a
  190 mg laser-powered aerial vehicle: The lightest wireless robot to fly,'' in
  \emph{Proc.~IEEE ICRA}, 2018.

\bibitem{martinec2007robust}
D.~Martinec and T.~Pajdla, ``Robust rotation and translation estimation in
  multiview reconstruction,'' in \emph{Proc.~IEEE CVPR}, 2007.

\bibitem{halperin2011tool}
D.~Halperin, W.~Hu, A.~Sheth, and D.~Wetherall, ``Tool release: Gathering
  802.11n traces with channel state information,'' \emph{ACM SIGCOMM Comput.
  Commun. Rev.}, 2011.

\bibitem{huang2011efficient}
J.~Huang, D.~Millman, M.~Quigley, D.~Stavens, S.~Thrun, and A.~Aggarwal,
  ``{Efficient, generalized indoor WiFi GraphSLAM},'' in \emph{Proc.~IEEE
  ICRA}, 2011.

\bibitem{li2016csi}
B.~Li, S.~Zhang, and S.~Shen, ``{CSI-based WiFi-inertial state estimation},''
  in \emph{Proc.~IEEE MFI}, 2016.

\bibitem{shen2015tightly}
S.~Shen, N.~Michael, and V.~Kumar, ``Tightly-coupled monocular visual-inertial
  fusion for autonomous flight of rotorcraft {MAVs},'' in \emph{Proc.~IEEE
  ICRA}, 2015.

\bibitem{dube2017online}
R.~Dub{\'e}, A.~Gawel, H.~Sommer, J.~Nieto, R.~Siegwart, and C.~Cadena, ``{An
  online multi-robot SLAM system for 3D LiDARs},'' in \emph{Proc.~IEEE IROS},
  2017.

\bibitem{hess2016real}
W.~Hess, D.~Kohler, H.~Rapp, and D.~Andor, ``{Real-time loop closure in 2D
  LiDAR SLAM},'' in \emph{Proc.~IEEE ICRA}, 2016.

\bibitem{6641260}
H.~Nyqvist and F.~Gustafsson, ``A high-performance tracking system based on
  camera and imu,'' in \emph{Proc.~IEEE FUSION}, 2013.

\bibitem{leutenegger2015keyframe}
S.~Leutenegger, S.~Lynen, M.~Bosse, R.~Siegwart, and P.~Furgale,
  ``Keyframe-based visual-inertial odometry using nonlinear optimization,''
  \emph{Int. J. Robotics Res.}, vol.~34, no.~3, pp. 314--334, 2015.

\bibitem{forster2015rss}
D.~F. S.~D. Forster~C, Carlone~L, ``Imu preintegration on manifold for
  efficient visual-inertial maximum-a-posteriori estimation,'' in
  \emph{Proc.~RSS}, 2015.

\bibitem{lupton2012visual}
T.~Lupton and S.~Sukkarieh, ``Visual-inertial-aided navigation for high-dynamic
  motion in built environments without initial conditions,'' \emph{IEEE Trans.
  Robot.}, vol.~28, no.~1, pp. 61--76, 2012.

\bibitem{kotaru2017position}
M.~Kotaru and S.~Katti, ``Position tracking for virtual reality using commodity
  wifi,'' in \emph{Proc.~IEEE CVPR}, 2017.

\bibitem{faria2005modeling}
D.~B. Faria \emph{et~al.}, ``Modeling signal attenuation in ieee 802.11
  wireless lans-vol. 1,'' \emph{Computer Science Department, Stanford
  University}, vol.~1, 2005.

\bibitem{mikhail2019flyzone}
A.~Mikhail, A.~Djordjevic, F.~Lui, and L.~Mottola, ``{FlyZone}: A testbed for
  experimenting with aerial drone applications,'' in \emph{Proc.~ACM MobiSys},
  2019.

\bibitem{crazyflie}
``Crazyflie 2.1,'' \url{https://www.bitcraze.io/crazyflie-2-1/}, {O}nline;
  accessed 20-December-2019.

\bibitem{palossi201964mw}
D.~Palossi, A.~Loquercio, F.~Conti, E.~Flamand, D.~Scaramuzza, and L.~Benini,
  ``A 64mw dnn-based visual navigation engine for autonomous nano-drones,''
  \emph{IEEE Internet of Things Journal}, 2019.

\end{thebibliography}

\end{document}